%% file: main.tex
\pdfoutput=1

\documentclass[11pt]{article}

\usepackage[preprint]{acl}

\usepackage{times}
\usepackage{latexsym}
\usepackage{booktabs} 
\usepackage{graphicx} 
\usepackage{lipsum}   
\usepackage{caption}  
\usepackage{geometry} 
\usepackage{algorithm}
\usepackage{algorithmicx}
\usepackage{algpseudocode}
\usepackage{subcaption} 
\usepackage{multirow}

\usepackage[T1]{fontenc}

\usepackage[utf8]{inputenc}

\usepackage{microtype}

\usepackage{inconsolata}

\usepackage{graphicx}

\title{\uqeval: Efficient Confidence Evaluation in NLG with Gold-Standard Correctness Labels}

\author{Xiaoou Liu*$^{1}$, Zhen Lin*, Longchao Da$^{1}$, Chacha Chen$^{2}$, Shubhendu Trivedi, Hua Wei$^{1}$\\
  ${^1}$ Arizona State University\\
  ${^2}$ University of Chicago\\
  Correspondence: \texttt{xiaoouli@asu.edu} \\
}

\input{defs}
\begin{document}
\maketitle
\begin{abstract}
Large Language Models (LLMs) require robust confidence estimation, particularly in critical domains like healthcare and law where unreliable outputs can lead to significant consequences. 
Despite much recent work in confidence estimation, current evaluation frameworks rely on \textit{correctness functions}---various heuristics that are often noisy, expensive, and possibly introduce systematic biases. 
These methodological weaknesses tend to distort evaluation metrics and thus the comparative ranking of confidence measures.
We introduce \uqeval, an evaluation framework for assessing confidence measures in Natural Language Generation (NLG) that eliminates dependence on an explicit correctness function by leveraging gold-standard correctness labels from multiple-choice datasets.
\uqeval enables systematic comparison of both internal state-based white-box (e.g. logit-based) and consistency-based black-box confidence measures, providing a unified evaluation methodology across different approaches.
Through extensive experiments on multiple LLMs and widely used QA datasets, we report that \uqeval provides efficient and more reliable assessments of confidence estimation methods than existing approaches.


\end{abstract}

\input{sections/1intro-hua}

\input{sections/2related_work}

\input{sections/3preliminary}

\input{sections/4method}

\input{sections/5experiments}

\input{sections/6conclusion}

\bibliography{main}

\input{sections/7appendix}

\end{document}

%% file: defs.tex
\usepackage{amsfonts}
\usepackage{amsmath}
\usepackage{amstext}
\usepackage{xcolor}         
\usepackage[capitalize]{cleveref}
\usepackage{breqn}
\usepackage{bbm}
\usepackage{amsthm}
\usepackage{amssymb}
\usepackage{mathtools}
\usepackage{xspace}
\usepackage{enumitem}
\usepackage[hang,flushmargin]{footmisc}

\newcommand{\predSeq}{\mathbf{s}}
\newcommand{\xInput}{\mathbf{x}}

\newcommand{\PredDist}{\mathcal{P}(S;\xInput,\mathcal{M})}

\newcommand{\uqeval}{\texttt{MCQA-Eval}\xspace}

\newcommand{\baselinePTrue}{\texttt{P(true)}\xspace}
\newcommand{\baselineSAR}{\texttt{TokenSAR}\xspace}
\newcommand{\baselineNLLNorm}{\texttt{Perplexity}\xspace}
\newcommand{\baselineSL}{\texttt{SL}\xspace}
\newcommand{\baselineCSLNext}{\texttt{CSL-Next}\xspace}
\newcommand{\baselineCSL}{\texttt{CSL}\xspace}
\newcommand{\baselineEcc}{\texttt{Ecc}\xspace}
\newcommand{\baselineDegree}{\texttt{Deg}\xspace}

\newcommand{\LMLLaMATwoName}{LLaMA2-7B\xspace}
\newcommand{\LMLLaMAThreeName}{LLaMA3-8B\xspace}
\newcommand{\phiName}{Phi4-14B\xspace}
\newcommand{\qwenName}{Qwen2.5-32b\xspace}

\newcommand{\acc}{f}

%% file: sections/1intro-hua.tex
\section{Introduction}\label{sec:intro}

Large Language Models (LLMs) demonstrate strong performance across natural language processing tasks, yet their architectural complexity and limited interpretability can produce unreliable outputs. 
This presents significant challenges in critical domains such as healthcare, where output errors carry serious consequences. 
Confidence estimation methods have emerged to quantify output reliability. 
The field connects closely with uncertainty quantification in natural language generation, as both address output trustworthiness. 
Current approaches divide into consistency-based methods, which analyze agreement across multiple outputs, and internal-states methods that leverage model-specific features like output probabilities.
Despite advances in these approaches, developing robust evaluation frameworks remains a central challenge.

Current evaluation frameworks for NLG confidence measures rely on correctness labels to compute metrics such as AUROC and AUARC. 
These frameworks follow a three-step process: generating model predictions, labeling correctness via a function $f(\cdot)$, and calculating metrics. 
This label-dependent approach faces several constraints. While human evaluation provides reliable correctness ground truth, it cannot scale to large datasets. 
Metrics based on reference matching, such as BLEU and ROUGE, fail to recognize semantically equivalent responses phrased differently.
LLM-based evaluators offer greater capability but remain noisy and may introduce systematic biases, such as favoring responses generated by themselves or similar LMs~\cite{panickssery2024llm}, or preferring longer responses~\cite{lin2022truthfulqa}.
Moreover, running such evaluators could be expensive.



Flaws in the correctness function $f(\cdot)$ propagate through the evaluation pipeline, affecting metrics like AUROC. This sensitivity becomes particularly problematic when comparing confidence estimation methods with similar performance. Such limitations underscore the need for evaluation frameworks that establish correctness more reliably.


In this paper, we propose \uqeval, a simple, efficient yet effective evaluation framework that eliminates the dependence on unreliable correctness functions.
\textit{The key insight is to leverage multiple-choice question-answering (QA) datasets, which inherently provide gold-standard answer choices at no cost.} 
With these definitive labels, our framework bypasses the ambiguity of determining correctness via correctness function $f(\cdot)$ and ensures an objective assessment of confidence estimation methods. 
Rather than replacing existing evaluation pipelines, our framework complements them, offering an additional lens to assess the discriminative power of confidence estimation methods.
\cref{fig:pipeline} shows how our proposal (green) and the existing evaluation pipeline (blue) differ, yet complement each other.
Our contributions are summarized as follows:
\begin{itemize}[leftmargin=*,nosep]
    \item We demonstrate that commonly used evaluation methods for NLG confidence measures are sensitive to noise in correctness labels, which can lead to misleading conclusions about evaluation metrics and rankings of different confidence estimation approaches.
    \item We propose a simple yet effective method that utilizes multiple-choice QA datasets to evaluate confidence measures, supporting both internal-states-based white-box and consistency-based black-box methods.
    \item Extensive experiments across recent LLMs and QA datasets verify that \uqeval produces stable evaluations broadly consistent with existing methods, while eliminating the need for expensive correctness functions.
\end{itemize}

%% file: sections/2related_work.tex
\section{Related Work}\label{sec:related}



\paragraph{Confidence Estimation}
Confidence estimation is fundamental to machine learning, providing mechanisms to assess model reliability and guide decision-making across tasks.
Early confidence estimation research concentrated on classification settings, where confidence scores enabled Selective Classification~\cite{geifman2017selective,el2010foundations,feng2022towards}—allowing models to abstain from low-quality predictions. 
The rapid advancement of NLG and LLMs has brought renewed attention to confidence estimation. 
While NLG poses unique challenges due to semantic invariance and vast output spaces~\cite{kuhn2023semantic}, recent works have advanced the field by measuring similarities among sampled responses~\cite{lin2024generating} and deriving measures from LMs' internal states~\cite{malininuncertainty,CSL,azaria-mitchell-2023-internal}.

A related aspect is calibration. While extensively considered in classification~\cite{ICML2020_MixNMatch_KDEEval,kull2019beyond,ICML2021_MetaCal}, it has received lass attention in NLG.
Since the distribution of confidence scores could vary significantly across different methods due to their underlying principles~\cite{geng2023survey, da2024llm}, calibrated confidence measures align better with human intuition for probabilities and are more interpretable~\cite{ICML2017_Guo,cosmides1996humans}.
While this paper focuses on evaluating confidence estimation methods, the same framework could be applied to evaluate future NLG calibration methods.
We demonstrate this by including results using common calibration metrics like Expected Calibration Error (ECE).

\paragraph{Evaluation of Confidence Measures}
While confidence estimation has received considerable attention, the evaluation of confidence measures remains under-explored. 
Many evaluation methods have been adapted from the classification literature, including Expected Calibration Error (ECE)~\cite{ICML2017_Guo,xiong2024can} and Area Under the Receiver Operating Characteristic Curve (AUROC)~\cite{kuhn2023semantic}. 
These metrics assess the relationship between confidence scores and prediction accuracy, typically requiring high-quality correctness labels for the evaluated responses.

However, obtaining reliable correctness labels in NLG is challenging due to factors such as semantic variability and ambiguity in open-ended tasks~\cite{novikova2017we}. 
Unlike classification where correctness is well-defined, NLG correctness is often determined through human annotation, LLM-based judges, or similarity-based comparisons between the generated and reference answers. 
These approaches are costly and often unreliable, as correctness judgments can be subjective and inconsistent~\cite{gatt2018survey}.

Recent works have attempted to mitigate these limitations. 
To allow for non-binary correctness measures, Rank Calibration Error (RCE)~\cite{RCEhuang2024} and AUARC~\cite{auarc,lin2024generating} were introduced, both of which leverage continuous correctness scores. 
Other approaches focus on improving correctness scores themselves. 
For example, \citet{CSL} aggregates predictions from multiple LLM-based judges and takes a consensus to enhance reliability.

Unlike these methods, our proposed framework completely circumvents the need for correctness labels, making it more robust and scalable for evaluating confidence measures in NLG. 


\paragraph{Applications of Confidence Measures}
Confidence measures play a crucial role in several downstream research areas in NLG, particularly in conformalized NLG and selective generation or generation with abstention. 
Stemming from Conformal Prediction~\cite{papadopoulos2007conformal}, in the context of NLG, conformalized methods typically aim to create a set of generation that satisfies a particular user-defined quality goal (e.g. ``correct answers'')~\cite{quach2023conformal,gui2024conformal,lee2024selective,yadkori2024mitigating}, or providing factual guarantees basing on parts of the generation~\cite{cherian2024large,mohri2024}.
Selective generation or generation with abstention, on the other hand, deals with broader considerations that involve refraining from generating if the confidence score is low, with goals like improving the accuracy on the non-rejected portion~\cite{MCConf-pmlr-v239-ren23a,cole2023selectively}.
Good confidence measures that can distinguish high and low-quality generations are key ingredients to all these research directions, and our paper aims to provide a better evaluation framework for researchers to identify such confidence measures.

%% file: sections/3preliminary.tex
\section{Confidence Estimation for NLG}\label{sec:prelim}

First, we establish notation and introduce relevant definitions.
Let $\mathcal{M}$ be a language model, $\xInput\in\Sigma^*$ be an input prompt, and $\predSeq=\mathcal{M}(\xInput)\in\Sigma^*$ be the output. 
$\Sigma$ denotes the vocabulary, which includes tokens from modern tokenizers or natural language symbols like alphabet letters. 
For free-form NLG datasets, we typically have reference answers $A={a_1,\ldots,a_m}$ alongside $\xInput$. A \textit{confidence estimation method} is a function that assigns a confidence score to model output $\predSeq$ given input $\xInput$. 
Formally, a confidence measure is defined as:
\begin{equation}
    C_{\mathcal{M}}: (\xInput, \predSeq) \in \Sigma^*\times\Sigma^*  \mapsto \mathbb{R},
\end{equation}
where $C_{\mathcal{M}}(\xInput,\predSeq)$ represents the confidence score of $\predSeq$. This notation accounts for both model-agnostic and model-specific confidence measures.




\subsection{Confidence Estimation Methods}
\label{confidence_methods}

Existing confidence estimation methods can be broadly divided into two categories: Consistency-based black-box methods and internal state-based white-box methods\footnote{We consider logits as an internal states here.}.


\textbf{Black-Box Methods} leverage response consistency across LLM generations~\cite{lin2024generating,manakul-etal-2023-selfcheckgpt}. 
Higher consistency among generated responses indicates higher confidence in $\predSeq$. 
These methods first compute pairwise response similarities, then derive confidence from the similarity matrix.
For similarity computation, existing methods use
Jaccard similarity, NLI models~\cite{he2021deberta}, and BERTScore~\cite{Zhang2020BERTScore} for similarity computation.


\textbf{White-Box Methods} use the internal states of LLMs—including logit distributions and token-level probabilities—to estimate confidence. 
Recent research has adopted sequence likelihood~\cite{CSL}, which computes confidence from the probability of the complete generated response.
\baselineNLLNorm~\cite{vashurin2024benchmarking} extends this by normalizing for response length via average sequence likelihood. 
Recent refinements weigh tokens differently: \baselineSAR~\cite{duan-etal-2024-shifting} uses NLI for token importance, while Contextualized Sequence Likelihood (\baselineCSL, and its variant\baselineCSLNext)~\cite{CSL} weighs using attention values. 
Other approaches train probes on LLM internal activations and embeddings~\cite{ren2023outofdistribution,azaria-mitchell-2023-internal,li2023inferencetime}. 
Furthermore, the verbalized confidence (\baselinePTrue)~\cite{xiong2024can} elicits explicit ``True'' or ``False'' predictions. 
While this is technically possible by taking the frequency of ``True'' among multiple sampled generations, in practice it is typically implemented by computing from the logits. Note that uncertainty quantification in NLG is a closely related research direction, yet differs in a key way: uncertainty characterizes the predictive distribution rather than a specific $\predSeq$. For more details of this distinction, see~\citet{lin2024generating}.

\begin{figure*}[h]
    \centering
    \includegraphics[width=\textwidth]{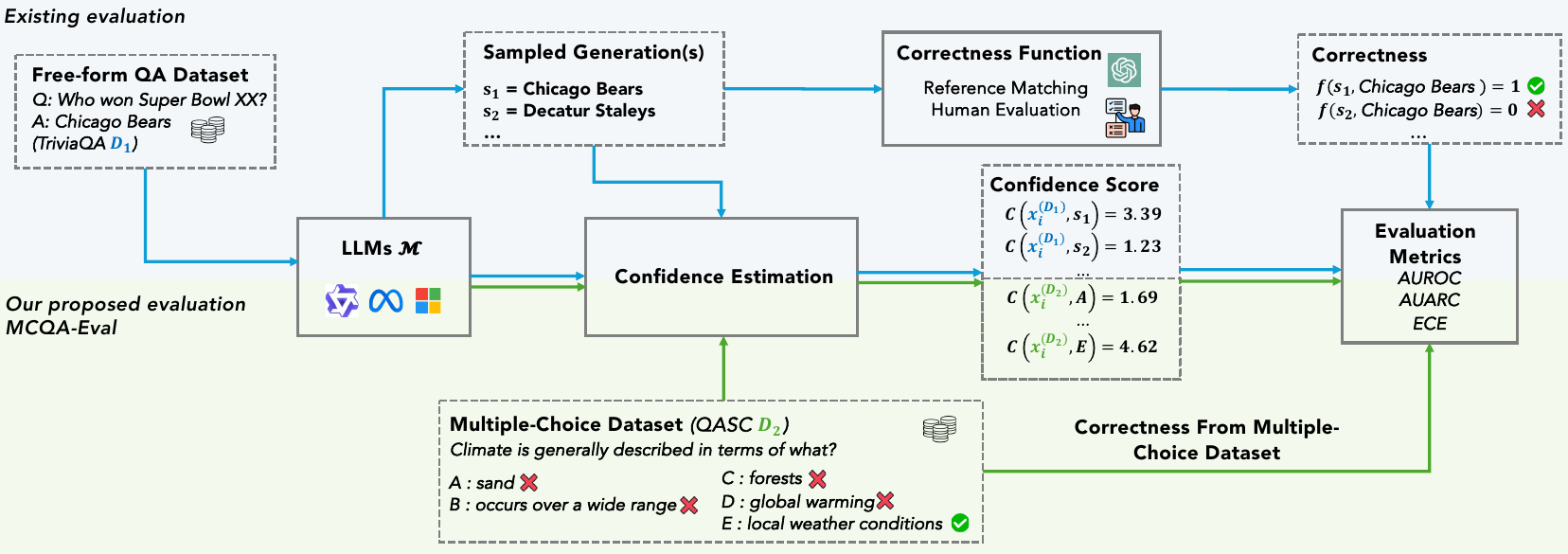}
    \caption{
    Illustration of the existing evaluation framework (blue) vs our proposed \uqeval (green).
    Unlike existing frameworks, we avoid the costly and unreliable correctness function module by using multiple-choice datasets.
    This requires slight modification to the confidence estimation steps, which is elaborated in \cref{sec:method}.
    }
    \label{fig:pipeline}
\end{figure*}


\subsection{Existing Evaluation Methods}\label{sec:prelim:old_eval}

Intuitively, a higher confidence score should correlate with the quality of model generation $\predSeq$ and its correctness relative to input $\xInput$. 
This assumption underpins selective classification, confidence scoring, and uncertainty quantification. 
In selective classification, also termed prediction with a rejection option, models abstain from low-confidence predictions, thereby reducing error rates while maximizing coverage~\cite{JMLR:v24:21-0048,geifman2017selective}. 
In other words, confidence measures guide selection towards predictions that are likely to be correct.


This idea extends naturally to NLG, where confidence measures are used to guide selective generation or generation with abstention.
Assuming a given \textit{correctness function}~\cite{RCEhuang2024} $\acc(\predSeq;\xInput)\in\{0,1\}$, which tells us whether a response is good or correct\footnote{This could sometimes be relaxed to have a continuous range of $\mathbb{R}$, instead of $\{0,1\}$, but certain evaluation metrics such as AUROC require binary correctness labels.}, several evaluation metrics are used to assess confidence measures for NLG:
\begin{itemize}[leftmargin=*, nosep]
    \item Area Under the Receiver Operating Characteristic Curve (AUROC): 
    \begin{equation}
        \int_{-\infty}^{\infty} \text{TPR}(t) \, d\text{FPR}(t),
    \end{equation}
    where $\text{TPR}(t)$ ($\text{FPR}(t)$) is the true (false) positive rate comparing $\mathbbm{1}\{C(\predSeq)>t\}$ and $\acc(\predSeq)$, the correctness of $\predSeq$.
    AUROC measures how well the confidence scores distinguish between correct and incorrect responses.
    
    \item Area Under the Accuracy-Rejection Curves (AUARC)~\cite{auarc}: 
    \begin{equation}
        \int_{-\infty}^{\infty} \text{Accuracy}(t) \, d\text{Coverage}(t),
    \end{equation}
    where $\text{Accuracy}(t)=\mathbb{E}\{\acc(\predSeq)|C(\predSeq)>t\}$ and $\text{Coverage}(t) = \mathbb{P}\{C(\predSeq)>t\}$.
    A refinement of AUROC designed for abstention-based settings, it evaluates the accuracy averaged across different coverage level (i.e. proportion of accepted predictions) when rejecting low-confidence predictions.
    
    \item Expected Calibration Error (ECE)~\cite{ICML2017_Guo}: 
    
    \begin{equation}
        \mathbb{E}\Big[|\mathbb{E}[\acc(\predSeq)|C(\predSeq)] - C(\predSeq)|\Big].
    \end{equation}
    ECE quantifies the alignment between predicted confidence scores and actual correctness probabilities.

    \item Rank-Calibratoin Error (RCE)~\cite{RCEhuang2024}: 
    {\small
    \begin{equation}
        \mathbb{E}_C\Big[|\mathbb{P}_{C'}\{reg(C')\geq reg(C)\} - \mathbb{P}_{C'}\{C'\leq C\}|\Big]
    \end{equation}
    }
    where $reg(c)$ is a regression function for $\mathbb{E}[\acc|C=c]$ and $C'$ and $C$ are the confidence values of two independent responses.
    Unlike ECE, which cannot be directly applied to confidence measures that have not been calibrated in the frequency space, RCE directly assesses calibration in the ranking space, and is more generally applicable.
    
\end{itemize}

While these evaluation metrics are widely used in classification tasks, they all rely on a \textbf{correctness function} $\acc(\predSeq)$ to decide if a generation $\predSeq$ is correct. 
However, in NLG, correctness is inherently difficult to determine, unless $\predSeq$ exactly matches one of the reference answers, which is rare except for simple tasks.
Currently, correctness is often assessed using human evaluation or similarity-based methods:
\paragraph{Human Evaluation} This remains arguably the most reliable approach. 
    Human evaluation is either used on smaller datasets~\cite{MCConf-pmlr-v239-ren23a} or to validate automated correctness functions~\cite{kuhn2023semantic,lin2024generating,CSL}, but is expensive and unscalable for large-scale dataset evaluation.
    
\paragraph{Similarity-Based Methods} In practice, correctness is often approximated by computing the similarity between $\predSeq$ and the reference answers $A$, in the form of $sim(\predSeq,A)$ or $sim(\predSeq,A|\xInput)$. 
    To accommodate metrics like AUROC, a threshold $\tau$ is applied to convert such similarity to $\{0,1\}$:
     \begin{equation}
        f(\predSeq, \xInput) =
        \begin{cases} 
          1, & \text{if } sim(\predSeq, A) > \tau \\ 
          0, & \text{otherwise}.
        \end{cases}\label{eq:correctness}
    \end{equation}
    Specifically, there are two common approaches for computing similarity.
    \textbf{Reference Matching} relies on lexical-based similarity metrics such as ROUGE and BLEU~\cite{hu2024unveiling,aynetdinov2024semscore,kuhn2023semantic}, which often fail to recognize semantically equivalent answers which are phrased differently.
    \textbf{LLM Judgment} uses a LLM as an evaluator~\cite{MCConf-pmlr-v239-ren23a,li2024generation,tan2024judgebench} and is more flexible. 
    However, such methods are computationally expensive and are still not fully reliable. 
    Recent studies indicate that machine-based correctness evaluation sometimes only has an accuracy of ~85\% on popular datasets~\cite{kuhn2023semantic,CSL}.




\subsection{Limitations of Existing Methods}
\label{subsecton:limitations}


Flaws in the correctness function inevitably affect downstream evaluation metrics such as AUROC and thus our conclusions about different confidence measures. 
In this section, we illustrate the limitations of current confidence evaluation methods from two angles: the impact of threshold sensitivity and the inherent noise of similarity measures.

\paragraph{Case Study 1: Threshold Sensitivity}

\begin{figure}[t]
  \includegraphics[width=1\columnwidth]{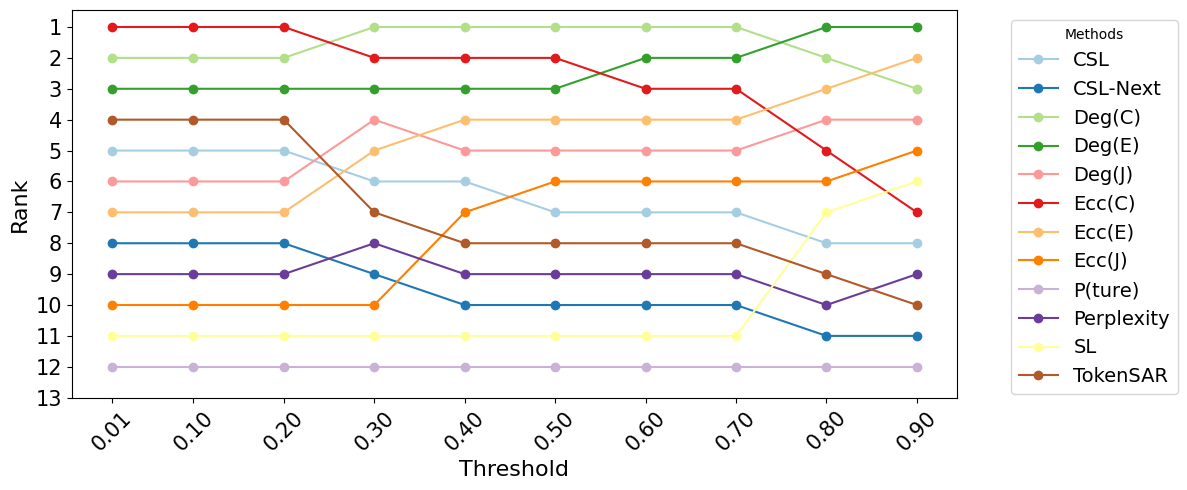}
  \caption{The AUROC ranking of black-box confidence measures (on LLaMA2-13B and CoQA) is sensitive to the threshold \( \tau \).
  }
  \label{fig:threshold}
\end{figure}

A common limitation of current practices is the need for a predefined threshold $\tau$ to convert similarity scores into binary correctness labels, as described in \cref{eq:correctness}. 
The choice of $\tau$ could thus impact the final evaluation metric.
To illustrate this, we vary the threshold for CoQA~\cite{CoQA} results from \citet{lin2024generating}, while keeping all other settings constant.
In their work, the threshold was manually set to $\tau=0.7$. 
However, \cref{fig:threshold} suggests that $\baselineEcc(C)$, for example, could either rank at the top or the bottom depending on $\tau$.

\begin{figure}[t]
  \includegraphics[width=\columnwidth]{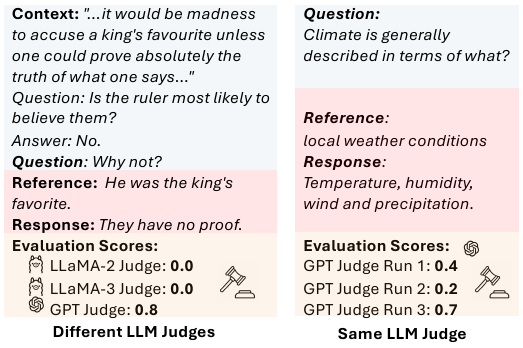}
  \caption{
  Using LLM judges as $\acc$, while flexible, still has inherent noise.
  Different LLMs disagree on whether a response is correct (left).
  Even the same LLM (GPT, right) could deliver different opinions simply due to the randomness in generation.
  }
  \label{fig:inconsis}
\end{figure}

\paragraph{Case Study 2: Similarity Noise} 
Correctness labels, whether derived from human evaluation, LLM-based scoring, or reference matching, are inherently noisy. 
For instance, within LLM-based judgments, correctness labels can fluctuate due to factors such as prompt variations and how the LLM judges were designed and trained. 
Echoing prior observations, \cref{fig:inconsis} shows examples where LLM judgments could either differ between different LLM judges or between different calls to the same LLM judge.
\citet{CSL} proposes to set the correctness function $\acc$ as the consensus of multiple LLMs, which improves the reliability of the correctness of responses LLMs agree on.
However, simply ignoring the disagreement could also introduce systematic selection bias.


To systematically analyze this effect, we simulate correctness label noise with Gaussian noise and analyze its effects.
We modify the correctness function as:
\begin{equation}
\small
    \tilde{\acc}(\predSeq; \xInput) = Sigmoid(logit(\acc(\predSeq; \xInput)) + \epsilon), \quad \epsilon \sim \mathcal{N}(0, \sigma^2).\label{eq:gaussian_noise}
\end{equation}
As shown in~\cref{tab:noise_rank}, increasing noise levels can lead to significant instability in ranking different confidence measures.
Note that our simulation likely \textit{underestimates} the issue, because the noise in \cref{eq:gaussian_noise} is unbiased and does not reflect systematic bias that may favor certain confidence measures~\cite{lin2022truthfulqa}.


\begin{table}[t]
\centering
\small
\resizebox{\columnwidth}{!}{ 
\begin{tabular}{@{}c|cccccc@{}}
\toprule
Ranking & 1 & 2 & 3 & 4 & 5 & 6  \\
\midrule
Original & Deg(C) & Deg(E) & Ecc(E) & Deg(J) & Ecc(C) & Ecc(J) \\
Noisy & Deg(C) & Deg(E) & Deg(J) & Ecc(E) & Ecc(J) & Ecc(C) \\
\bottomrule
\end{tabular}}
\vspace{-2mm}
\caption{Ranking of uncertainty quantification methods before and after noise.}
\label{tab:noise_rank}
\end{table}

While it might be theoretically possible to estimate the noise level and its propagation to errors on metrics like AUROC, this requires strong assumptions (e.g. \cref{eq:gaussian_noise}), extensive human evaluation, and replication across LLM judges and datasets. 
This fragility in existing evaluation methods motivates our framework, which eliminates dependence on uncertain correctness functions.



%% file: sections/4method.tex
\section{ \uqeval: A framework for Assessing Confidence Estimation}\label{sec:method}
At a high level, existing evaluation frameworks
for $C_{\mathcal{M}}$ includes three main steps (blue path in \cref{fig:pipeline}):
\begin{enumerate}[nosep]
    \item Generate $\predSeq$ from $\mathcal{M}$ given the input $\xInput_i$.
    \item Determine the correctness label of $\predSeq$ using the function $\acc(\cdot,\xInput)$.
    \item Compute evaluation metrics such as AUROC. A higher metric value indicates that $C_{\mathcal{M}}$ is a ``better'' confidence estimation.
\end{enumerate}
The main limitation of this general pipeline lies in $\acc$ in step 2. 
Existing evaluation frameworks all implicitly assume step 1---that the confidence measure $C_{\mathcal{M}}$ must apply to generated sequences $\predSeq$. 
While this might hold for consistency-based uncertainty measures, where response divergence indicates uncertainty, it does not extend to confidence measures. 
In other words, we could relax step 1 in order to improve step 2.


\textit{Our main proposal in this paper is to adapt multiple-choice datasets to evaluate confidence measures designed for free-form NLG.}
Unlike free-form NLG datasets, multiple-choice datasets provide inherent correctness values for options, eliminating the need for an explicit correctness function. 
If we simply ``pretend'' that these options are free-form generations from the base LM, we can directly evaluate the confidence measure quality. 
As \cref{fig:pipeline} shows, the approach differs from existing evaluation pipelines only in applying confidence estimation methods to multiple-choice options.

\begin{figure}[t]
  \includegraphics[width=\columnwidth]{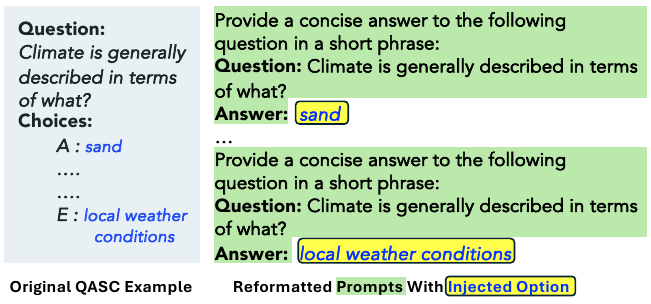}
  \caption{
  We reformat each option from the multiple-choice question (left), by injecting the \smash{\colorbox{yellow!40}{{{\color{blue}option}}}} to a free-form QA \smash{\colorbox{green!40}{prompt}}.
  One could typically apply any confidence estimation method by treating this \smash{\colorbox{yellow!40}{{{\color{blue}option}}}} as if it was generated by the base LM.
  For black-box confidence measures that require additional responses, we only feed the \smash{\colorbox{green!40}{prompt}} to the base LM.
  }
  \label{fig:qasc_example}
\vspace{-3mm}
\end{figure}

Consider the QASC~\cite{khot2020qasc} dataset as an example,
each problem comes with a question $\xInput$ and a few choices, $o_1,\ldots,o_K$. 
Unlike what such datasets were designed for, we re-format the input prompt as a free-form NLG question, as illustrated in \cref{fig:qasc_example}, as if the base LLM generated each option itself, in different runs.
In what follows, we first explain explain slight nuances in applying internal state-based white-box confidence measures as well as consistency-based black-box ones. 

\textbf{Logit or Internal State-Based Measures} typically examine the internals of a LM when it generates a particular response.
The nature of the free-form generation task allows us to simply plug-in the option $o_i$ into the corresponding location of the prompt, and extract similar information that allows us to evaluate the confidence\footnote{In fact, this was the practice to compute \baselineSL for actual generations. For example, \url{https://github.com/lorenzkuhn/semantic_uncertainty/blob/main/code/get_likelihoods.py} and \url{https://huggingface.co/docs/transformers/perplexity}.}.



\textbf{Consistency-based Confidence Measures}
Unlike logit-based or internal-state-based measures, consistency-based confidence measures typically rely on an estimate of the predictive distribution, denoted as $\PredDist$, and any response that is closer to the center of the distribution (in the ``semantic space'') is considered to be of higher confidence. 
Consider methods from~\citet{lin2024generating} as an example. To preserve the integrity of the predictive distribution, we first sample $n$ responses from $\PredDist$ as usual, and then iteratively include one option $o_i$ at a time to compute its associated confidence score~\cite{rivera-etal-2024-combining,manakul-etal-2023-selfcheckgpt}. 
\cref{alg:confidence_score} outlines this process.

\renewcommand{\algorithmicrequire}{\textbf{Input:}}
\renewcommand{\algorithmicensure}{\textbf{Output:}}

\begin{algorithm}[t]
\small
\caption{Consistency-based Confidence Estimation for Any Sequences}
\label{alg:confidence_score}
\begin{algorithmic}[1]
    \Require $\xInput$, $\mathcal{M}$, candidate sequences $A = \{a_1, \dots, a_K\}$
    \Ensure $\{C_{\mathcal{M}}(\xInput,a_1), \dots,C_{\mathcal{M}}(\xInput,a_K)\}$ 
    
    \State Generate $S = \{\predSeq_1, \dots, \predSeq_{n}\}$ using $\mathcal{M}$ for question $\xInput$
    \State Compute pairwise similarity matrix $M$ of $S$.
    
    \For{each $a_i \in A$}
        \State Compute a new similarity matrix $M_i$ of $S\cup\{a_i\}$, reusing $M$. 
        \State Compute confidence score $C_{\mathcal{M}}(\xInput,a_i)$ using $M_i$. 
    \EndFor

    \State \Return $\{C_{\mathcal{M}}(\xInput,a_1), \dots,C_{\mathcal{M}}(\xInput,a_K)\}$ 
\end{algorithmic}
\end{algorithm}



\paragraph{Remarks}
Our proposal relaxes step 1 at the beginning of this section, allowing for $\predSeq^*=o_i$ not sampled from $\PredDist$.
This is not to be misunderstood as a proposal to \textit{replace} the current pipeline (\cref{sec:prelim:old_eval})---rather, it is \textit{complementary}.
The rationale is that if a good confidence measure predicts the correctness well, it should perform well in \textit{both} evaluation frameworks.
In fact, any $o_i\in\Sigma^*$ that does not violate the generation configuration, has a non-zero probability to be sampled from $\PredDist$, and a robust confidence measure should be expected to model it well.


%% file: sections/5experiments.tex
\section{Experiments}\label{sec:experiments}

We demonstrate the advantages of our proposed evaluation framework through comprehensive experiments on multiple LLMs and various confidence estimation methods.

\subsection{Experimental Setup}

\paragraph{Base LLMs}
Our experiments use four popular open-source LLMs: \LMLLaMATwoName ~\cite{touvron2023llama2}, \LMLLaMAThreeName, \phiName~\cite{abdin2024phi}, and \qwenName~\cite{yang2024qwen2}. 
These models were specifically pretrained on question-answering tasks, which minimizes irrelevant responses.
We include various model sizes for a comprehensive analysis.

\paragraph{Datasets}

We select five multiple-choice datasets with varying levels of complexity from different domains, including CommonSenseQA(C-QA)~\cite{talmor-etal-2019-commonsenseqa}, Question Answering via Sentence Composition (QASC)~\cite{khot2020qasc}, MedQA~\cite{jin2021disease}, RACE-m, and RACE-h~\cite{lai2017race}. 
Each dataset consists of independent questions with a set of answer options, where exactly one option is correct. 
Evaluating different LLMs on datasets from different domains and diverse levels of difficulty allows for a more comprehensive assessment of model performance across a wide range of scenarios. 
\cref{tab:datasets} provides an overview of these datasets and the number of questions we use, with detailed descriptions in \cref{sec:datasetDes}.
The exact prompt formulation for each dataset is provided in \cref{sec:appendix_prompt}. 

\begin{table}[t]
\centering
\scriptsize
\begin{tabular}{@{}lcccl@{}}
\toprule
\textbf{Dataset} & \textbf{Size} & \textbf{Options} & \textbf{Domain} & \textbf{Difficulty} \\ 
\midrule
C-QA & 1221 & 5 & Commonsense & Easy \\ 
QASC & 926 & 8 & Commonsense & Medium \\ 
MedQA & 1000 & 5 & Medical & Hard \\ 
RACE-M & 1000 & 4 & Reading Comprehension & Medium \\ 
RACE-H & 1000 & 4 & Reading Comprehension & Hard \\ 
\bottomrule
\end{tabular}
\vspace{-1mm}
\caption{Overview of datasets used in this paper.}
\vspace{-5mm}
\label{tab:datasets}
\end{table}

\paragraph{Confidence Estimation Methods}

We compare six black-box and six white-box methods.
The selected methods represent commonly used confidence estimation baselines.
The six \textbf{black-box} measures evaluated in our experiments are:
\begin{itemize}[leftmargin=*, nosep]
    \item \baselineDegree(J), \baselineDegree(E), \baselineDegree(C): These compute the similarity matrix using Jaccard Similarity, NLI entailment and NLI contradiction, respectively. 
    The confidence score is then derived from the degree matrix.
    
    \item \baselineEcc(J), \baselineEcc(E), \baselineEcc(C): 
      The similarity matrix is obtained using the same method as above, but the confidence score is derived from the embeddings derived from the graph Laplacian.
\end{itemize}

Unlike black-box methods, \textbf{white-box} measures directly use the multiple-choice options as evaluation responses. 
We implement six white-box confidence estimation baselines, as introduced in Section~\ref{confidence_methods}: \baselineSL, \baselineNLLNorm, \baselineSAR, \baselineCSL, its variant \baselineCSLNext, and \baselinePTrue.

\paragraph{Metrics}
\label{sec:metrics}
Following previous works,
we use AUROC as our primary metric\footnote{Responses sampled for black-box methods' are excluded from AUROC calculations due to uncertain correctness.}. Our framework can also be applied to evaluate confidence calibration. We include additional results in \cref{sec:full_results}, reporting RCE and calibration ECE metrics. 

Additional details of our experiment can be found in \cref{appendix:sec:exp_imp}.


\subsection{Experimental Findings}
This section summarizes our main experimental findings, with detailed results in \cref{sec:full_results}.

\paragraph{Comparison With Existing Evaluation Methods}
We first compare our evaluation method with the existing pipeline (Baseline) using the QASC dataset. 
For the baseline, we use gpt-4o-mini to obtain correctness labels (by comparing the generation with the correct option).
\cref{tab:uq_ranking} shows that varying the threshold significantly impacts the ranking of both black-box and white-box confidence estimation methods.
Additionally, querying gpt-4o-mini for correctness labels across $926 \times 20$ responses takes approximately 2.5 hours. 
The cost (both economical and time-wise) would be much higher for more advanced LLM judges, or longer prompts from datasets with a ``context'' (such as CoQA~\cite{CoQA}), making large-scale evaluations difficult. 

On the other hand, \uqeval aligns with baseline ranking at $\tau=0.9$.
While it is unclear in this case which $\tau$ reflects the ``most reliable'' ranking, this experiment suggests that \uqeval's conclusion is consistent with existing pipeline.
However, unlike the Baseline, it does not require the costly correctness function(s), thereby reducing computational costs and enabling scalable evaluation.


\begin{table}[t]
\small
\centering
\resizebox{\columnwidth}{!}{
\begin{tabular}{@{}clllllll@{}}
\toprule
 &  & \multicolumn{6}{c}{\textbf{Ranking}} \\ 
\cmidrule(lr){3-8}
 & $\tau$ & \textbf{1} & \textbf{2} & \textbf{3} & \textbf{4} & \textbf{5} & \textbf{6} \\ 
 \midrule\midrule
 &        & \multicolumn{6}{c}{\textbf{Black-box}} \\ 
\cmidrule(lr){3-8}
\multirow{5}{*}{\textbf{Baseline}} 
    
    
    & 0.5 & Deg(C) & Deg(E) & Ecc(E) & Ecc(C) & Deg(J) & Ecc(J) \\ 
    & 0.7 & Deg(C) & Deg(E) & Ecc(C) & Ecc(E) & Deg(J) & Ecc(J) \\ 
    & 0.9 & Ecc(E) & Deg(E) & Deg(C) & Deg(J) & Ecc(J) & Ecc(C) \\ 
\midrule
\textbf{\uqeval} & N/A & Ecc(E) & Deg(E) & Deg(J) & Deg(C) & Ecc(J) & Ecc(C) \\ 
\midrule\midrule
 &        & \multicolumn{6}{c}{\textbf{White-box}} \\ \cmidrule(lr){3-8}
\multirow{5}{*}{\textbf{Baseline}} 
   
    
    & 0.5 & TokenSAR & Perplexity & SL & CSL & CSL-Next & P(true) \\ 
    & 0.7 & TokenSAR & Perplexity & CSL & CSL-Next & SL & P(true) \\ 
    & 0.9 & SL & TokenSAR & Perplexity & CSL & CSL-Next & P(true) \\ 
\midrule
\textbf{\uqeval} & N/A & SL & TokenSAR & Perplexity & CSL & CSL-Next & P(true) \\

\bottomrule
\end{tabular}}
\vspace{-1mm}
\caption{
We analyze how existing LLM-based evaluation methods rank black-box and white-box approaches by varying $\tau$ from 0.9 to 0.5.
\uqeval aligns with the rankings at $\tau=0.9$, yet requires no overhead for the correctness function.}
\vspace{-2mm}
\label{tab:uq_ranking}
\end{table}

\begin{figure}[t]
  \centering
  \begin{subfigure}[b]{\columnwidth}
    \centering
    \includegraphics[width=\columnwidth]{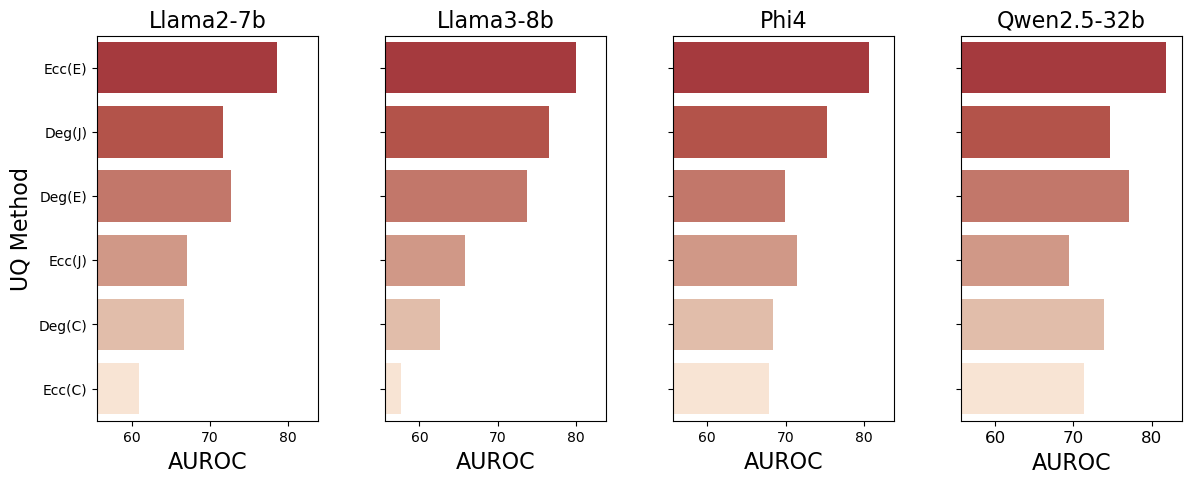}
    \caption{AUROC of different black-box methods.}
    \label{fig:cqa_blackbox}
  \end{subfigure}
  \begin{subfigure}[b]{\columnwidth}
    \centering
    \includegraphics[width=\columnwidth]{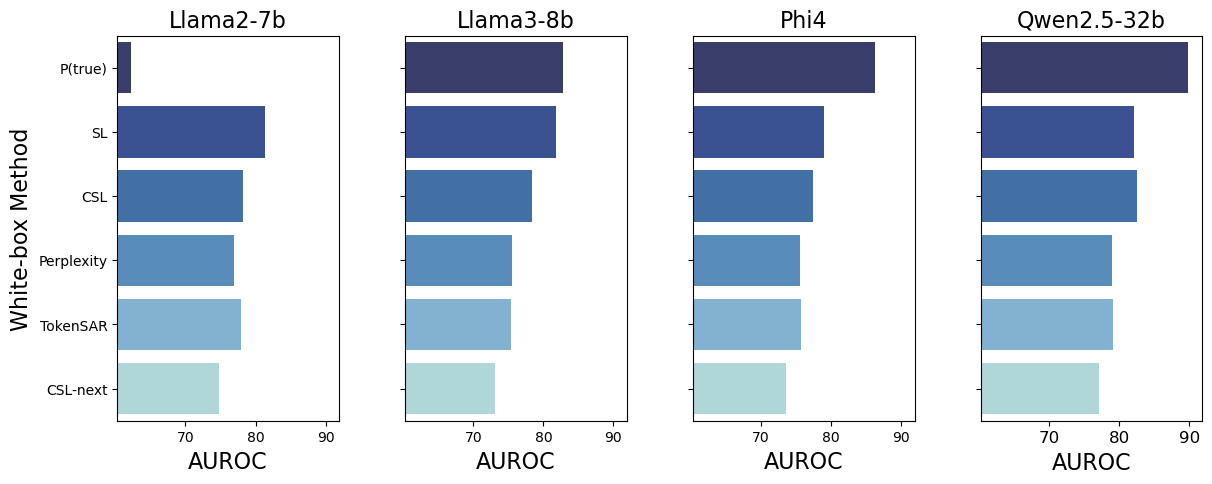}
    \caption{AUROC of different white-box methods.}
    \label{fig:another_dataset}
  \end{subfigure}

  \caption{(a) and (b) show the performance of 4 different LLMs and 12 different confidence estimation methods on the C-QA dataset. A higher AUROC indicates better performance.}
  \label{fig:llm_perspective}
\end{figure}

\paragraph{Comparison Across LLMs }
We compare confidence measures across LLMs on the same dataset via  \uqeval.
As shown in \cref{fig:llm_perspective} (with additional results available in the \cref{sec:full_results}), larger LLMs tend to achieve better performance across different confidence estimation methods, reflecting their broader pretraining exposure. 
The relatively ranking of various confidence measures stay mostly stable.
Interestingly, unlike some prior results~\cite{lin2024generating,vashurin2024benchmarking}, \baselinePTrue performs very well except for Llama2-7b. 
We hypothesize that this is due to improvement in recent LLMs' abilities in general, which is similar to the conjecture in \cite{vashurin2024benchmarking}.
This hypothesis is partially supported by the fact that \baselinePTrue performs increasingly well as the base LM becomes more sophisticated.


\begin{figure}[t]
  \includegraphics[width=\columnwidth]{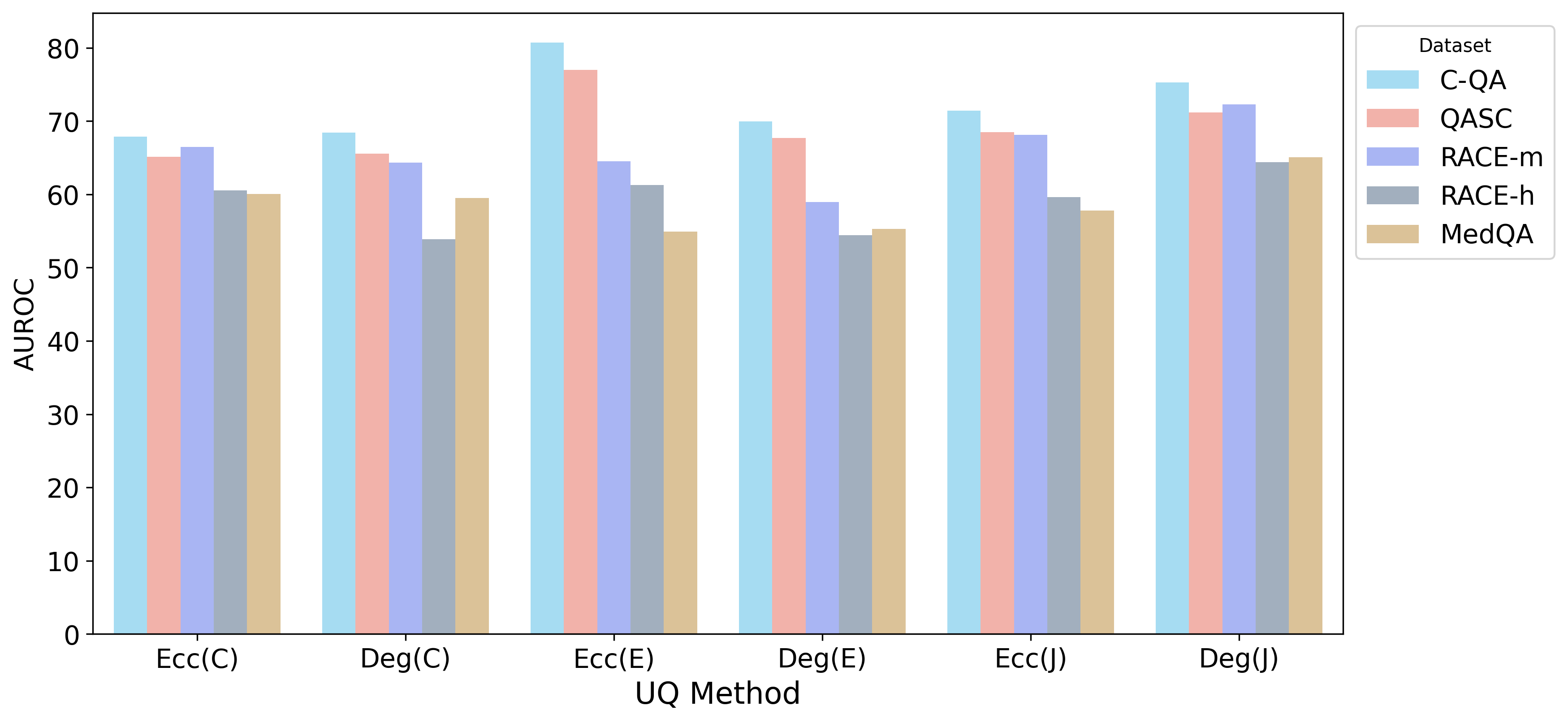}
  \caption{The performance of Phi-4 using black-box methods across different datasets.}
  \label{fig:dataset_perspective}
  \vspace{-5mm}
\end{figure}

\paragraph{Comparison Across Datasets }
We compare confidence measures across datasets of varying difficulty with \uqeval. 
\cref{fig:dataset_perspective} illustrates these results for \phiName. In general, different confidence measures exhibit larger performance gap on simpler datasets such as C-QA, and smaller on more professional datasets like MedQA.
The general poor performance on harder datasets could be attributed to limited capability of the base LM (in generating additional responses for black-box measures, or in supplying the base logits for white-box measures.
For black-box measures, similarity metrics like NLI and Jaccard may also provide limited distinguishing power.
We recommend selecting datasets with difficulty levels that align with the capabilities of the language model, making performance differences more discernible.



%% file: sections/6conclusion.tex
\section{Conclusion}\label{sec:conclusion}

In this paper, we propose \uqeval, a simple framework using multiple-choice QA datasets to evaluate confidence measures for natural language generation.
We first highlight the unreliability of widely used \textit{correctness functions} in existing evaluation frameworks.
To address this, we propose an alternative approach that reformulates multiple-choice questions into a free-form QA prompt, enabling a more efficient evaluation with higher-quality correctness labels. 
Experiments across diverse datasets and state-of-the-art LLMs demonstrate that \uqeval produces consistent results aligned with prior research findings, while eliminating dependence on costly correctness functions.


\section*{Limitations}
While our proposed evaluation framework avoids the use of correctness functions, offering speed and reliability, it also has its limitations.
As noted in \cref{sec:method}, \uqeval should not serve as the \emph{only} evaluation method.
Bypassing response generation provides no guarantee that injected options resemble what would otherwise be generated by the base LM.
If the goal is to evaluate confidence measures \textit{for a specific LM and its generation}, then this generation step by definition should not be skipped.
Further, certain trained confidence measures (e.g. linear probes on the LM's internal states) might not generalize as well to injected options, and may perform systematically worse in \uqeval than in the current framework (although one may argue that generalizability should be part of the evaluation to begin with).
Finally, \uqeval currently only applies to confidence measures, but we do not see a straightforward adaption to uncertainty measures. 
We hope future research could continue to improve the evaluation of confidence, and potentially, of uncertainty measures.

%% file: sections/7appendix.tex
\cleardoublepage

\appendix

\def \TabNotation{
\begin{table}[]
\centering
\resizebox{!}{0.18\textwidth}{
\begin{tabular}{@{}lc@{}}
\toprule
\textbf{Notation} & \textbf{Explanation} \\ 
\midrule
$\mathcal{M}$ & Language model \\ 
$\Sigma$ & Vocabulary \\
$\xInput$ & Input prompt \\ 
$\predSeq$ & Output responds \\ 
$A= \{ a_i,...a_m\}$ & Set of reference answers \\ 
$C_{\mathcal{M}}(\xInput,\predSeq)$ & Confidence score of $\predSeq$ \\ 
$f(\predSeq,\xInput)$ & Correctness function \\
$sim(\predSeq,A)$ & Similarity score \\
$\tau$ & LLM predefined threshold \\
$n$ & Number of open-form responses \\
$o_i$ & Options in QA dataset \\
$K$ & Number of options \\
\bottomrule
\end{tabular}}
\vspace{-1mm}
\caption{The notation used in this paper}
\vspace{-5mm}
\label{tab:notations}
\end{table}
}


\section{Experiments Details}\label{appendix:sec:exp_imp}

\subsection{Dataset Description}\label{sec:datasetDes}

\begin{itemize}
    \item \textbf{C-QA} A multiple-choice dataset designed for commonsense question answering. Each question requires world knowledge and reasoning to determine the correct answer from 5 given choices. The dataset consists of 1,221 test questions.
    
    \item \textbf{QASC} A multiple-choice commonsense reasoning dataset with 8 answer choices per question. Compared to C-QA, QASC presents a higher level of difficulty. While the dataset was originally designed for multi-hop reasoning, our focus is not on evaluating the reasoning capabilities of LLMs. Therefore, we do not provide the supporting facts to the model and instead present only the question. For our experiments, we use the original validation set, which includes 926 questions.
    
    \item \textbf{MedQA} A multiple-choice dataset with 5 options for answers, specifically designed for medical QA. 
    It covers three languages: English, simplified Chinese, and traditional Chinese, and contains 12,723, 34,251, and 14,123 questions for the three languages, respectively.
    The questions are sourced from professional medical board exams, making this dataset particularly challenging due to its reliance on specialized medical knowledge. 
    For our experiments, we randomly selected the first 1,000 questions from the English dataset.
    
    \item \textbf{RACE-m and RACE-h} used in this paper are derived from the RACE (\textbf{R}e\textbf{A}ding \textbf{C}omprehension dataset from \textbf{E}xaminations) dataset, a large-scale machine reading comprehension dataset introduced by Lai et al~\cite{lai2017race}. 
    RACE comprises 27,933 passages and 97,867 questions collected from English examinations for Chinese students aged 12–18. 
    These datasets evaluate a model’s ability to comprehend complex passages and answer questions based on contextual reasoning. 
    Each question is accompanied by four answer choices, with only one correct option. 
    For our experiments, we randomly sampled 1,000 questions from the entire dataset using a fixed random seed of 42 to ensure reproducibility.
\end{itemize}

\subsection{Prompt Details}
\label{sec:appendix_prompt}
\begin{itemize}
    \item We use the following prompt to collect open-form responses for each of the 5 datasets separately.
    
\includegraphics[width=.9\columnwidth]{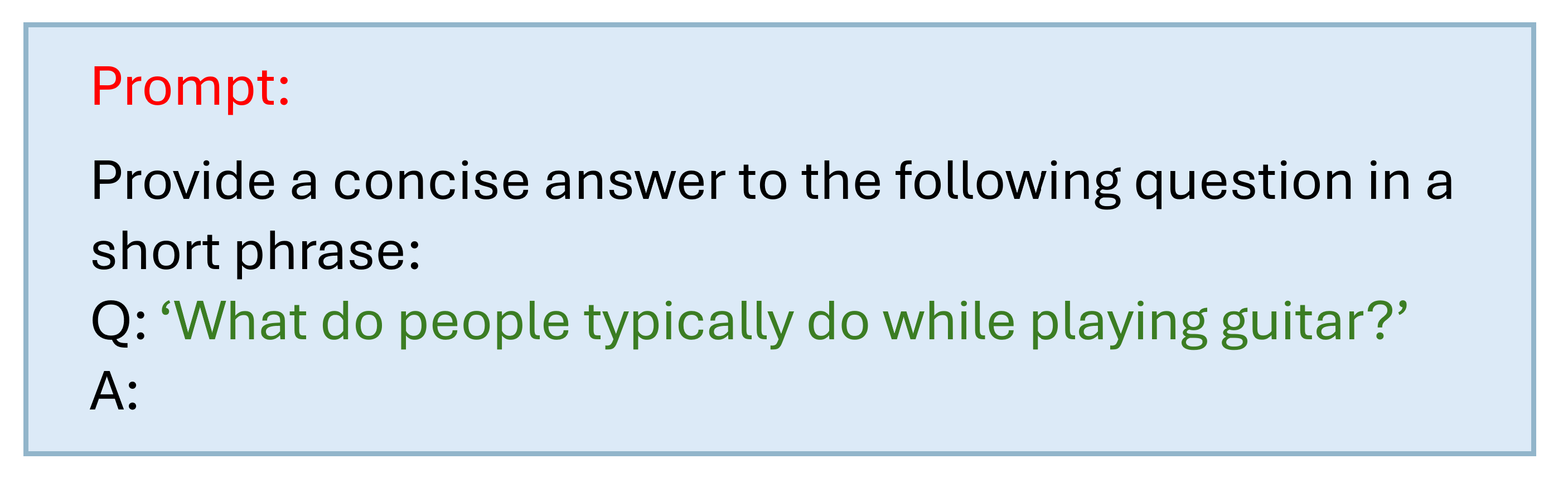}

    \item We use the following prompt to elicit P(True) confidence score.
    The ``Possible Answer'' is an option from the multiple-choice dataset.
    
\includegraphics[width=.9\columnwidth]{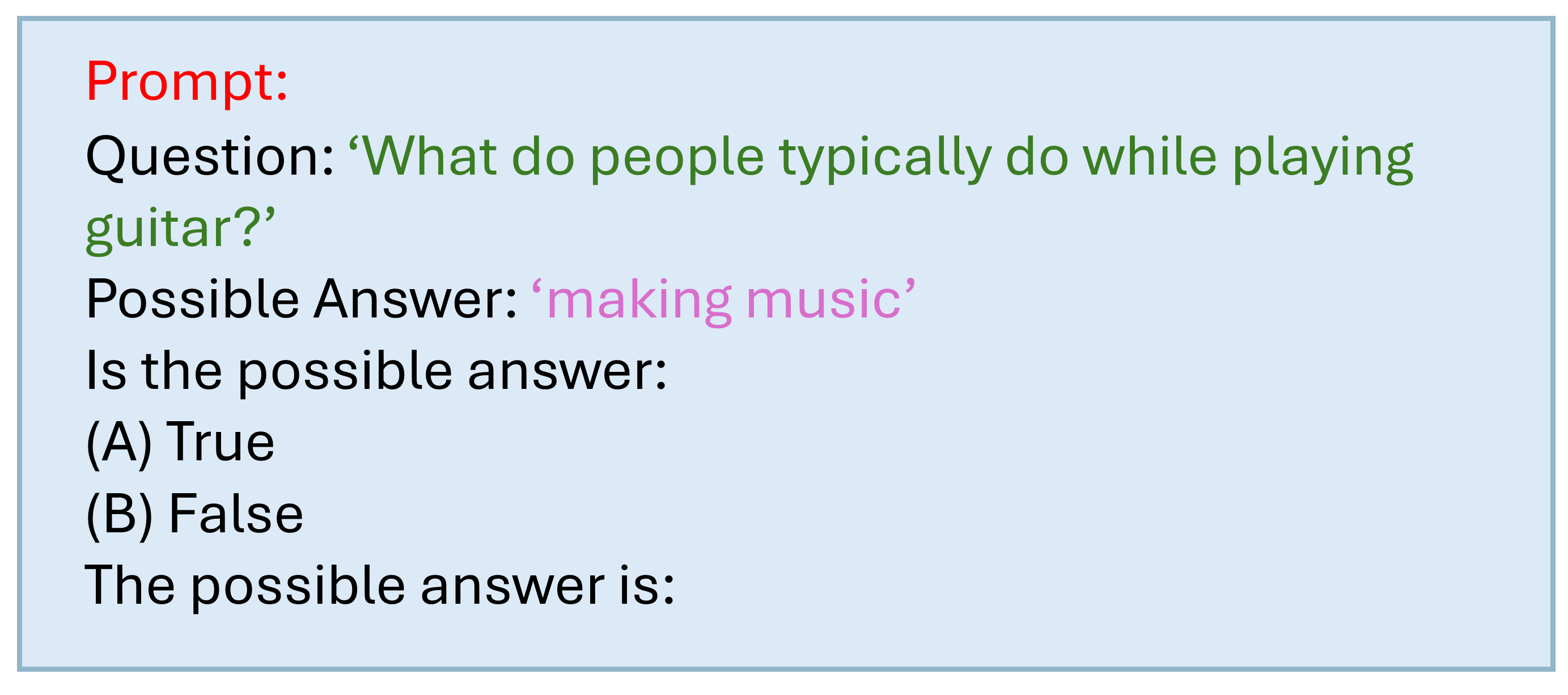}
\end{itemize}

\subsection{Computation Resources}
To efficiently process multiple queries, we used vLLM~\cite{kwon2023efficientvllm} for parallel inference.
All experiments were conducted on a Linux server running Ubuntu, equipped with an A100 80GB GPU.

\subsection{Response Generation }
For black-box methods, we mostly adopt the experimental configurations from~\citet{lin2024generating}. 
Sampling-based black-box confidence measures use $n=20$ open-form responses per question. 
The temperature settings for different LLMs are kept at their default values.

\section{Additional Experiments Results}\label{sec:full_results}

\subsection{Full Results of Different Evaluation Metrics}
In the main text, due to space constraints, we only show a subset of the AUROC results.
Here, \cref{appendix:tab:bb:auc,appendix:tab:wb:auc} show the AUROC and AUARC for black-box and white-box confidence measures, respectively. 
Similarly, \cref{appendix:tab:bb:calib,appendix:tab:wb:calib} present RCE and ECE results.
Note that all ECE are based on \textit{calibrated} confidence measures for fair comparisons, as some original confidence measures are not even constrained to $[0,1]$.
For the calibration step, we applied histogram binning method~\cite{KDD_HistogramBinning} on all methods.

\begin{table*}[h!]
\centering
\resizebox{\textwidth}{!}{%
\begin{tabular}{llcccccccccccc}
\toprule
\multirow{2}{*}{\textbf{Dataset}} & \multirow{2}{*}{\textbf{Model}} & \multicolumn{6}{c}{\textbf{AUROC $\Uparrow$}} & \multicolumn{6}{c}{\textbf{AUARC} $\Uparrow$} \\ 
\cmidrule(lr){3-8} \cmidrule(lr){9-14}
 &  & Ecc(C) & Deg(C) & Ecc(E) & Deg(E) & Ecc(J) & Deg(J) & Ecc(C) & Deg(C) & Ecc(E) & Deg(E) & Ecc(J) & Deg(J) \\ 
\midrule
\multirow{4}{*}{C-QA}
 & Llama2-7b   & 60.981 & 66.651 & 78.629 & 72.771 & 67.081 & 71.668 & 29.386 & 33.266 & 38.221 & 35.858 & 34.681 & 36.915 \\
 & Llama3-8b   & 57.590 & 62.592 & 80.004 & 73.734 & 65.886 & 76.583 &32.062 & 33.232 & 38.414 & 32.648 & 37.150 & 38.596\\
 & Phi4        & 67.879 & 68.413 & 80.712 & 69.976 & 71.447 & 75.278 & 32.123 & 31.596 & 19.294 & 30.032 & 28.570 & 24.739 \\
 & Qwen2.5-32b & 71.409 & 73.931 & 81.885 & 77.087 & 69.473 & 74.645 & 34.775 & 37.399 & 39.926 & 37.964 & 36.776 & 38.808 \\
\midrule
\multirow{4}{*}{QASC}
 & Llama2-7b   & 58.949 & 61.978 & 73.221 & 69.200 & 61.659 & 66.877 & 17.509 & 19.628 & 25.724 & 23.556 & 21.469 & 23.251 \\
 & Llama3-8b   & 55.121 & 55.446 & 74.912 & 72.033 & 64.124 & 72.657 & 15.785 & 15.952 & 25.199 & 24.163 & 23.198 & 25.786 \\
 & Phi4        & 65.100 & 65.553 & 76.980 & 67.692 & 68.496 & 71.209 & 20.297 & 21.063 & 26.740 & 21.422 & 24.067 & 24.308 \\
 & Qwen2.5-32b & 62.218 & 61.611 & 74.546 & 71.702 & 64.658 & 69.131 & 19.522 & 19.830 & 25.695 & 24.306 & 23.182 & 24.510  \\
\midrule
\multirow{4}{*}{MedQA}
 & Llama2-7b   & 53.683 & 54.129 & 52.076 & 52.963 & 53.137 & 53.778 & 21.956 & 23.105 & 21.160 & 22.863 & 23.454 & 23.371 \\
 & Llama3-8b   & 52.824 & 53.971 & 51.641 & 53.523 & 55.257 & 59.552 & 21.125 & 22.103 & 20.390 & 22.164 & 25.598 & 26.617 \\
 & Phi4        & 60.055 & 59.512 & 54.945 & 55.261 & 57.815 & 65.067 & 25.081 & 25.410 & 22.077 & 22.940 & 27.573 &29.201 \\
 & Qwen2.5-32b & 60.071 & 61.737 & 54.727 & 58.454 & 61.564 & 63.783 & 24.998 & 28.045 & 22.246 & 26.331 & 29.848 & 30.054 \\
\midrule
\multirow{4}{*}{RACE-m}
 & Llama2-7b  & 65.473 & 64.304 & 61.022 & 59.245 & 67.480 & 67.760 & 34.147 & 36.637 & 32.570 & 33.994 & 38.844 & 38.904 \\
 & Llama3-8b & 62.385 & 63.351 & 61.872 & 58.711 & 68.391 & 73.267 & 30.774 & 35.054 & 31.639 & 32.491 & 41.231 & 43.055 \\
 & Phi4     & 66.461 & 64.344 & 64.492 & 58.981 & 68.124 & 72.304 & 34.312 & 35.355 & 32.903 & 32.232 & 41.311 & 41.895  \\
 & Qwen2.5-32b & 65.425 & 67.627 & 60.268 & 61.309 & 75.420 & 75.746 & 34.393 & 37.409 & 32.092 & 34.850 & 44.281 & 44.585 \\
\midrule
\multirow{4}{*}{RACE-h}
 & Llama2-7b    & 58.991 & 53.597 & 57.178 & 54.037 & 59.300 & 59.856 & 34.147 & 36.637 & 32.570 & 33.994 & 38.844 & 38.904 \\
 & Llama3-8b  & 56.372 & 53.560 & 58.456 & 54.004 & 57.488 & 63.788 & 27.959 & 28.483 & 29.120 & 27.823 & 33.912 & 36.139 \\
 & Phi4    & 60.550 & 53.867 & 61.263 & 54.442 & 59.639 & 64.385 & 30.733 & 28.641 & 31.411 & 28.157 & 34.519 & 35.710    \\
 & Qwen2.5-32b   & 60.012 & 54.781 & 55.984 & 55.657 & 64.985 & 66.130 & 31.049 & 29.180 & 30.459 & 28.921 & 37.620 & 37.734 \\
\bottomrule
\end{tabular}%
}
\caption{AUROC and AUARC for black-box methods, across different models and datasets}
\label{appendix:tab:bb:auc}
\end{table*}

\begin{table*}[h!]
\centering
\resizebox{\textwidth}{!}{%
\begin{tabular}{llcccccccccccc}
\toprule
\multirow{2}{*}{\textbf{Dataset}} & \multirow{2}{*}{\textbf{Model}} & \multicolumn{6}{c}{\textbf{AUROC} $\Uparrow$} & \multicolumn{6}{c}{\textbf{AUARC} $\Uparrow$} \\ \cmidrule(lr){3-8} \cmidrule(lr){9-14}
 &  & P(true) & CSL & CSL-next & SL & Perplexity & TokenSAR & P(true) & CSL & CSL-next & SL & Perlexity & TokenSAR \\ \midrule
\multirow{4}{*}{C-QA}
 & Llama2-7b   & 62.278 & 78.253 & 74.799 & 81.390 & 76.958 & 77.888 &  28.401 & 38.231 & 36.213 & 40.178 & 37.579 & 37.450 \\
 & Llama3-8b   & 82.760 & 78.423 & 73.068 & 81.731 & 75.503 & 75.385 & 40.235 & 38.191 & 35.096 & 40.152 & 36.368 & 35.453  \\
 & Phi4        & 86.184 & 77.382 & 73.477 & 78.903 & 75.471 & 75.722 & 42.447 & 37.984 & 35.749 & 38.452 & 36.928 & 36.630  \\
 & Qwen2.5-32b & 89.892 & 82.486 & 77.087 & 82.143 & 78.964 & 79.064 & 45.449 & 40.802 & 38.003 & 40.596 & 38.674 & 38.215\\
 \midrule
\multirow{4}{*}{QASC}
 & Llama2-7b   & 66.198 & 77.535 & 76.053 & 79.589 & 77.637 & 77.696 &  19.815 & 25.986 & 25.494 & 27.632 & 26.324 & 25.921\\
 & Llama3-8b   & 86.069 & 77.970 & 73.090 & 80.718 & 74.531 & 75.006 &  30.127 & 26.215 & 24.251 & 28.253 & 24.442 & 24.308 \\
 & Phi4        & 84.478 & 77.556 & 74.596 & 78.661 & 75.678 & 76.222 & 29.977 & 26.068 & 25.246 & 27.064 & 25.463 &25.307 \\
 & Qwen2.5-32b & 88.998 & 79.324 & 73.895 & 78.598 & 74.485 & 75.175 & 32.992 & 26.810 & 24.608 & 27.387 & 24.069 & 23.992  \\
  \midrule
\multirow{4}{*}{MedQA}
 & Llama2-7b   & 54.660 & 55.144 & 55.852 & 54.766 & 55.766 & 55.703 & 22.414 & 24.437 & 24.888 & 24.246 & 24.848 & 24.795  \\
 & Llama3-8b   & 77.493 & 57.384 & 57.894 & 57.919 & 57.592 & 57.530 & 36.884 & 24.072 & 25.225 & 25.879 & 24.973 & 24.803 \\
 & Phi4 & 86.888 & 65.550 & 64.284 & 63.287 & 65.588 & 65.696 &42.615 & 31.671 & 31.050 & 30.888 &31.752 & 31.775  \\
 & Qwen2.5-32b & 80.131 & 63.264 & 63.712 & 63.109 & 62.564 & 62.164 &40.197 & 27.495 & 27.754 & 29.382 & 27.440 & 27.221  \\
  \midrule
\multirow{4}{*}{RACE-m}
 & Llama2-7b  & 63.965 & 69.194 & 70.819 & 67.568 & 71.823 & 71.984 & 35.543 & 38.429 & 39.404 & 38.870 & 40.030 & 40.133 \\
 & Llama3-8b   & 82.118 & 67.317 & 70.875 & 69.321 & 69.851 & 70.029 & 47.145 & 36.953 & 40.206 & 40.508 & 39.144 & 39.232 \\
 & Phi4        & 90.543 & 68.334 & 69.5354 & 68.8049 & 69.025 & 69.188 & 52.457 & 36.638 & 38.717 & 40.314 & 37.972 & 38.057 \\
 & Qwen2.5-32b  & 56.049 &67.294 & 69.102 & 73.267 & 69.147 & 69.279 & 29.283 & 34.913 & 36.873 & 42.373 & 36.220 & 36.318 \\
  \midrule
\multirow{4}{*}{RACE-h}
 & Llama2-7b   & 61.265 & 61.905 & 62.481 & 59.889 & 63.486 & 63.465 & 35.543 & 38.429 & 39.404 & 38.870 & 40.030 & 40.133 \\
 & Llama3-8b    & 79.466 & 60.775 & 63.868 & 61.253 & 64.134 & 64.146 & 44.910 & 31.300 & 34.086 & 33.463 & 33.973 & 33.974 \\
 & Phi4       & 87.172 & 62.253 & 62.680 & 60.178 & 63.391 & 63.383 &  50.250 & 32.395 & 33.484 & 33.243 & 33.547 & 33.537 \\
 & Qwen2.5-32b   & 52.811 & 61.837 & 64.047 & 63.555 & 64.050 & 64.024 & 27.605 & 31.279 & 32.714 & 34.462 & 32.462 & 32.458 \\
\bottomrule
\end{tabular}%
}
\caption{AUROC and AUARC for white-box methods, across different models and datasets}
\label{appendix:tab:wb:auc}
\end{table*}

\begin{table*}[t]
\centering
\resizebox{\textwidth}{!}{%
\begin{tabular}{llcccccccccccc}
\toprule
\multirow{2}{*}{\textbf{Dataset}} & \multirow{2}{*}{\textbf{Model}} & \multicolumn{6}{c}{\textbf{RCE}} & \multicolumn{6}{c}{\textbf{Calibration ECE}} \\ \cmidrule(lr){3-8} \cmidrule(lr){9-14}
 &  & Ecc(C) & Deg(C) & Ecc(E) & Deg(E) & Ecc(J) & Deg(J) & Ecc(C) & Deg(C) & Ecc(E) & Deg(E) & Ecc(J) & Deg(J) \\ \midrule
\multirow{4}{*}{C-QA} 
 & Llama2-7b    & 0.2857  & 0.143722 & 0.117486 & 0.084357 & 0.271789 & 0.198744 & 0.014457 & 0.064792 & 0.025161 & 0.009014 & 0.009546 & 0.031801 \\
 & Llama3-8b    & 0.28071 & 0.15255  & 0.06311  & 0.041246 & 0.362527 & 0.153761 & 0.013865 & 0.044074 & 0.031566 & 0.016865 & 0.008845 & 0.060919 \\
 & Phi4         & 0.18881 & 0.115068 & 0.067507 & 0.038771 & 0.225698 & 0.218135 & 0.017734 & 0.059135 & 0.040364 & 0.024237 & 0.019987 & 0.056875 \\
 & Qwen2.5-32b  & 0.16192 & 0.114378 & 0.080021 & 0.055613 & 0.278165 & 0.198222 & 0.0111   & 0.087857 & 0.043406 & 0.016647 & 0.014439 & 0.051092 \\
  \midrule
\multirow{4}{*}{QASC} 
 & Llama2-7b    & 0.25132 & 0.162559 & 0.193186 & 0.121908 & 0.331258 & 0.252667 & 0.013984 & 0.020481 & 0.019263 & 0.012321 & 0.003108 & 0.022164 \\
 & Llama3-8b    & 0.28697 & 0.231308 & 0.083146 & 0.057512 & 0.401264 & 0.230094 & 0.003117 & 0.005336 & 0.004844 & 0.009398 & 0.010951 & 0.022145 \\
 & Phi4         & 0.19064 & 0.104986 & 0.066258 & 0.063753 & 0.23061  & 0.225091 & 0.004181 & 0.015734 & 0.012447 & 0.01108  & 0.003271 & 0.026654 \\
 & Qwen2.5-32b  & 0.25004 & 0.142512 & 0.091264 & 0.084393 & 0.31938  & 0.272657 & 0.010503 & 0.020774 & 0.012144 & 0.009716 & 0.004127 & 0.023387 \\
  \midrule
\multirow{4}{*}{MedQA} 
 & Llama2-7b    & 0.19817 & 0.188788 & 0.231296 & 0.243174 & 0.263178 & 0.213793 & 0.005909 & 0.006271 & 0.006057 & 0.01008  & 0.007157 & 0.008915 \\
 & Llama3-8b    & 0.21067 & 0.190038 & 0.286932 & 0.194414 & 0.290058 & 0.146904 & 0.006035 & 0.006757 & 0.006424 & 0.006872 & 0.01166  & 0.007277 \\
 & phi4         & 0.09127 & 0.09877  & 0.208792 & 0.132527 & 0.308812 & 0.087518 & 0.008327 & 0.018021 & 0.0156   & 0.008231 & 0.020912 & 0.016443 \\
 & Qwen2.5-32b  & 0.09064 & 0.089393 & 0.194414 & 0.087518 & 0.234422 & 0.118149 & 0.006312 & 0.01598  & 0.011337 & 0.021417 & 0.014092 & 0.021119 \\
  \midrule
\multirow{4}{*}{RACE-m} 
 & Llama2-7b  & 0.09876 & 0.31881 & 0.17315 & 0.27630 & 0.14502 & 0.16065 & 0.04523 & 0.07009 & 0.01980 & 0.01965  & 0.00778 & 0.01433 \\
 & Llama3-8b  & 0.10252 & 0.32068 & 0.12877 & 0.27005 & 0.21254 & 0.04500 & 0.00939 & 0.08513 & 0.00962 & 0.04675 & 0.025705  & 0.03261 \\
 & phi4          & 0.06001 & 0.31756  & 0.11252 & 0.26817 & 0.150655 & 0.07501 & 0.01699 & 0.07599 & 0.03366   & 0.01936 & 0.016385 &  0.01542 \\
 & Qwen2.5-32b  & 0.19378 &  0.32756 &  0.18253 & 0.27505 & 0.09689 & 0.1187 & 0.024623 & 0.10445  & 0.02922 & 0.05540 & 0.01300 & 0.02171 \\
  \midrule
\multirow{4}{*}{RACE-h} 
 & Llama2-7b  & 0.12565 & 0.36069 & 0.22441 & 0.40383 & 0.29568 & 0.30881 & 0.01702 & 0.06116 & 0.01635 & 0.01577  & 0.020679 & 0.01569 \\
 & Llama3-8b   & 0.20316 & 0.37007 & 0.18816 & 0.42070 & 0.26192 & 0.05938 & 0.01754 & 0.06838 & 0.01672 & 0.02324 & 0.02597  & 0.02622\\
 & phi4        & 0.09751 & 0.36757  & 0.14627 & 0.38820 &  0.26880 & 0.15878 & 0.01928 & 0.06393 & 0.021709   & 0.02191 & 0.02294 & 0.02502 \\
 & Qwen2.5-32b  & 0.11564 & 0.35069 & 0.21441 & 0.35569 & 0.28505 & 0.205666 & 0.01679 & 0.06562  & 0.01794 &0.02833 & 0.015137 & 0.01438 \\[1ex]
\bottomrule
\end{tabular}%
}
\caption{RCE and (calibrated) ECE for black-box methods, across different models and datasets}
\label{appendix:tab:bb:calib}
\end{table*}

\begin{table*}[t]
\centering
\resizebox{\textwidth}{!}{%
\begin{tabular}{llcccccccccccc}
\toprule
\multirow{2}{*}{\textbf{Dataset}} & \multirow{2}{*}{\textbf{Model}} & \multicolumn{6}{c}{\textbf{RCE}} & \multicolumn{6}{c}{\textbf{Calibration ECE}} \\ \cmidrule(lr){3-8} \cmidrule(lr){9-14}
 &  & P(true) & CSL & CSL-next & SL & SL(norm) & TokenSAR & P(true) & CSL & CSL-next & SL & SL(norm) & TokenSAR \\ \midrule
\multirow{4}{*}{C-QA} 
 & Llama2-7b    & 0.084386 & 0.0506   & 0.041895  & 0.041267     & 0.038126 & 0.034997 & 0.0102   & 0.035637  & 0.041958  & 0.023881        & 0.04454 & 0.027278 \\
 & Llama3-8b    & 0.040614 & 0.03563  & 0.068102  & 0.031902     & 0.057489 & 0.038742 & 0.01871  & 0.034739  & 0.050008  & 0.022294        & 0.04291 & 0.026352 \\
 & Phi4         & 0.043731 & 0.04626  & 0.046892  & 0.043771     & 0.041858 & 0.03501  & 0.0583   & 0.034232  & 0.055943  & 0.019535        & 0.04302 & 0.030655 \\
 & Qwen2.5-32b  & 0.058105 & 0.02999  & 0.044359  & 0.032513     & 0.044363 & 0.059406 & 0.0369   & 0.022175  & 0.046935  & 0.021905        & 0.03671 & 0.021438 \\
  \midrule
\multirow{4}{*}{QASC} 
 & Llama2-7b    & 0.077448 & 0.04685  & 0.078796  & 0.051258     & 0.043136 & 0.045007 & 0.01119  & 0.024505  & 0.037871  & 0.023245        & 0.0326  & 0.023127 \\
 & Llama3-8b    & 0.030627 & 0.04811  & 0.117522  & 0.050664     & 0.08503  & 0.043753 & 0.00894  & 0.020665  & 0.038958  & 0.025687        & 0.03274 & 0.020785 \\
 & Phi4         & 0.082518 & 0.04437  & 0.116905  & 0.066942     & 0.088115 & 0.049376 & 0.02122  & 0.021401  & 0.0415    & 0.028242        & 0.02548 & 0.033083 \\
 & Qwen2.5-32b  & 0.11997  & 0.04878  & 0.062505  & 0.081237     & 0.073773 & 0.041861 & 0.03096  & 0.014358  & 0.040047  & 0.025111        & 0.02665 & 0.023483 \\
   \midrule
\multirow{4}{*}{MedQA} 
 & Llama2-7b    & 0.181911 & 0.19254  & 0.19879   & 0.191288     & 0.228796 & 0.238798 & 0.00606  & 0.015623  & 0.007533  & 0.00791         & 0.00669 & 0.007449 \\
 & Llama3-8b    & 0.028131 & 0.08939  & 0.121274  & 0.207542     & 0.163158 & 0.178161 & 0.0166   & 0.012721  & 0.008949  & 0.03            & 0.00861 & 0.010613 \\
 & phi4         & 0.05126  & 0.09127  & 0.115648  & 0.176285     & 0.119399 & 0.116273 & 0.02853  & 0.046391  & 0.05184   & 0.058272        & 0.05787 & 0.05535  \\
 & Qwen2.5-32b  & 0.078141 & 0.06126  & 0.07314   & 0.128151     & 0.088143 & 0.075015 & 0.03067  & 0.020881  & 0.033491  & 0.047763        & 0.03295 & 0.032673 \\
   \midrule
\multirow{4}{*}{RACE-m} 
 & Llama2-7b  & 0.16253 & 0.26130 & 0.22254 & 0.13502     & 0.24317 & 0.24567 & 0.00741 & 0.01935 & 0.03113 & 0.01820  & 0.061396 & 0.062452 \\
 & Llama3-8b    & 0.05938 & 0.18003 & 0.09814 & 0.12752     & 0.10252 & 0.12189 & 0.05006 & 0.05534 & 0.04303 & 0.04812        & 0.01986  & 0.02156 \\
 & phi4        & 0.09689 & 0.15753  & 0.09314 & 0.09689     & 0.13127 & 0.13565 & 0.02585 & 0.04808 &0.02775  & 0.032335        & 0.01727 & 0.01938 \\
 & Qwen2.5-32b  & 0.16940 & 0.17566 & 0.17691 & 0.17691     & 0.24567 & 0.25255 &0.00695& 0.04720  & 0.05091 &  0.07564        & 0.07986 &  0.08066 \\
   \midrule
\multirow{4}{*}{RACE-h} 
 & Llama2-7b & 0.17566 & 0.31818 & 0.32318 & 0.33256     & 0.31818 & 0.32256 & 0.01748 & 0.02600 & 0.01719 & 0.01613         & 0.021382 & 0.021339 \\
 & Llama3-8b   & 0.05563 & 0.22316 & 0.12189 & 0.15565     & 0.163782 & 0.149404 & 0.045399 & 0.01684 & 0.031577 & 0.034098        & 0.030134  & 0.030341 \\
 & phi4       & 0.08939 & 0.19566  & 0.150030 & 0.13315     & 0.19316 & 0.19566 & 0.019294 & 0.035576 &  0.02874   & 0.03037        & 0.02238 & 0.040637 \\
 & Qwen2.5-32b    & 0.24754 & 0.22254 & 0.21754 & 0.21316     & 0.29505 & 0.30006 & 0.016826 & 0.02004  &0.02105 & 0.022801        & 0.03156 & 0.04110 \\
\bottomrule
\end{tabular}%
}
\caption{RCE and (calibrated) ECE for white-box methods, across different models and datasets}
\label{appendix:tab:wb:calib}
\end{table*}

\subsection{Additional Visualizations for ROC Curves}
\begin{figure*}[htbp]
    \centering
    \begin{subfigure}[b]{0.45\textwidth}
        \centering
        \includegraphics[width=\textwidth]{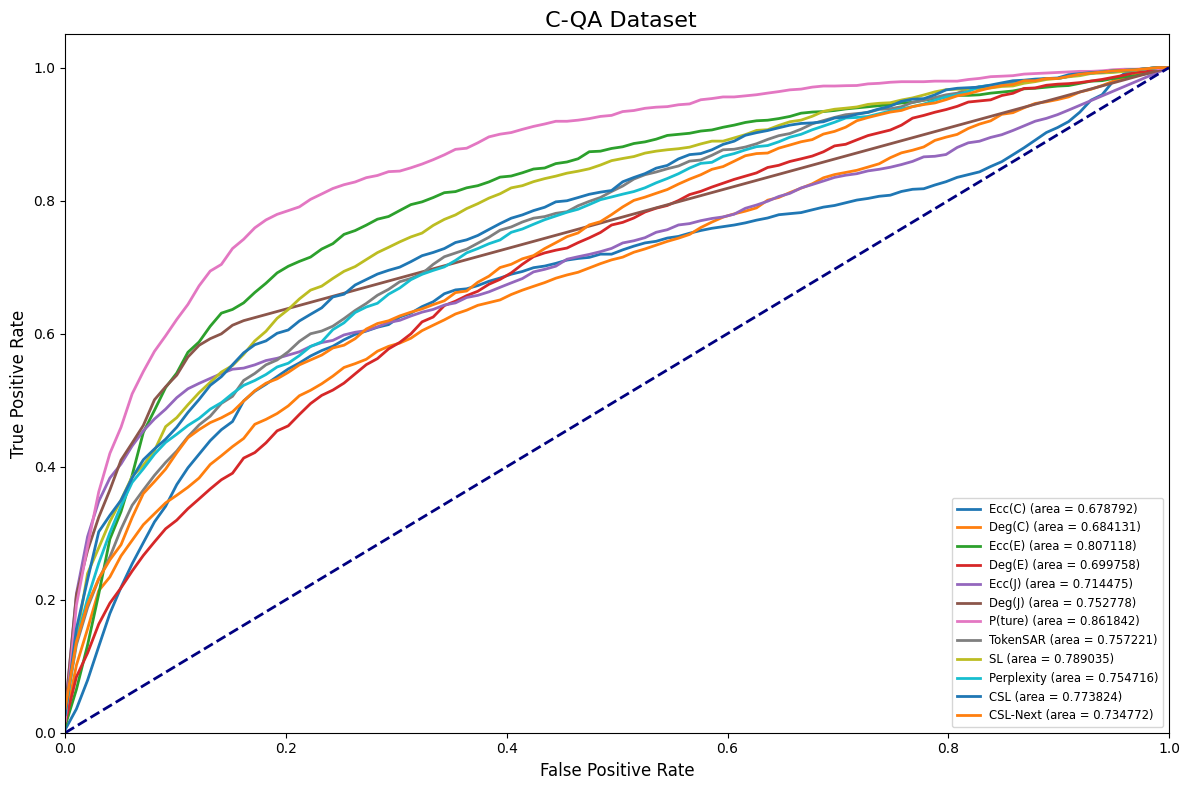}
        \caption{C-QA Dataset}
        \label{fig:subfig1}
    \end{subfigure}
    \hfill
    \begin{subfigure}[b]{0.45\textwidth}
        \centering
        \includegraphics[width=\textwidth]{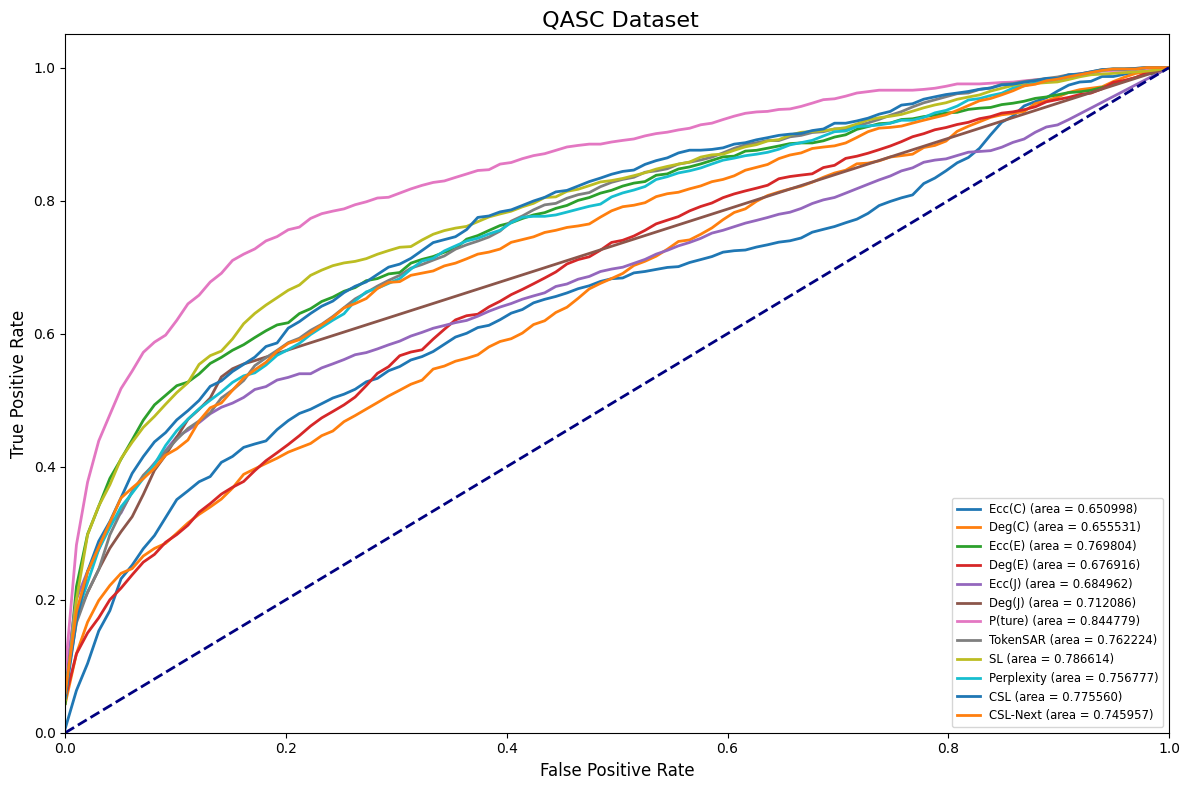}
        \caption{QASC Dataset}
        \label{fig:subfig2}
    \end{subfigure}

    \begin{subfigure}[b]{0.45\textwidth}
        \centering
        \includegraphics[width=\textwidth]{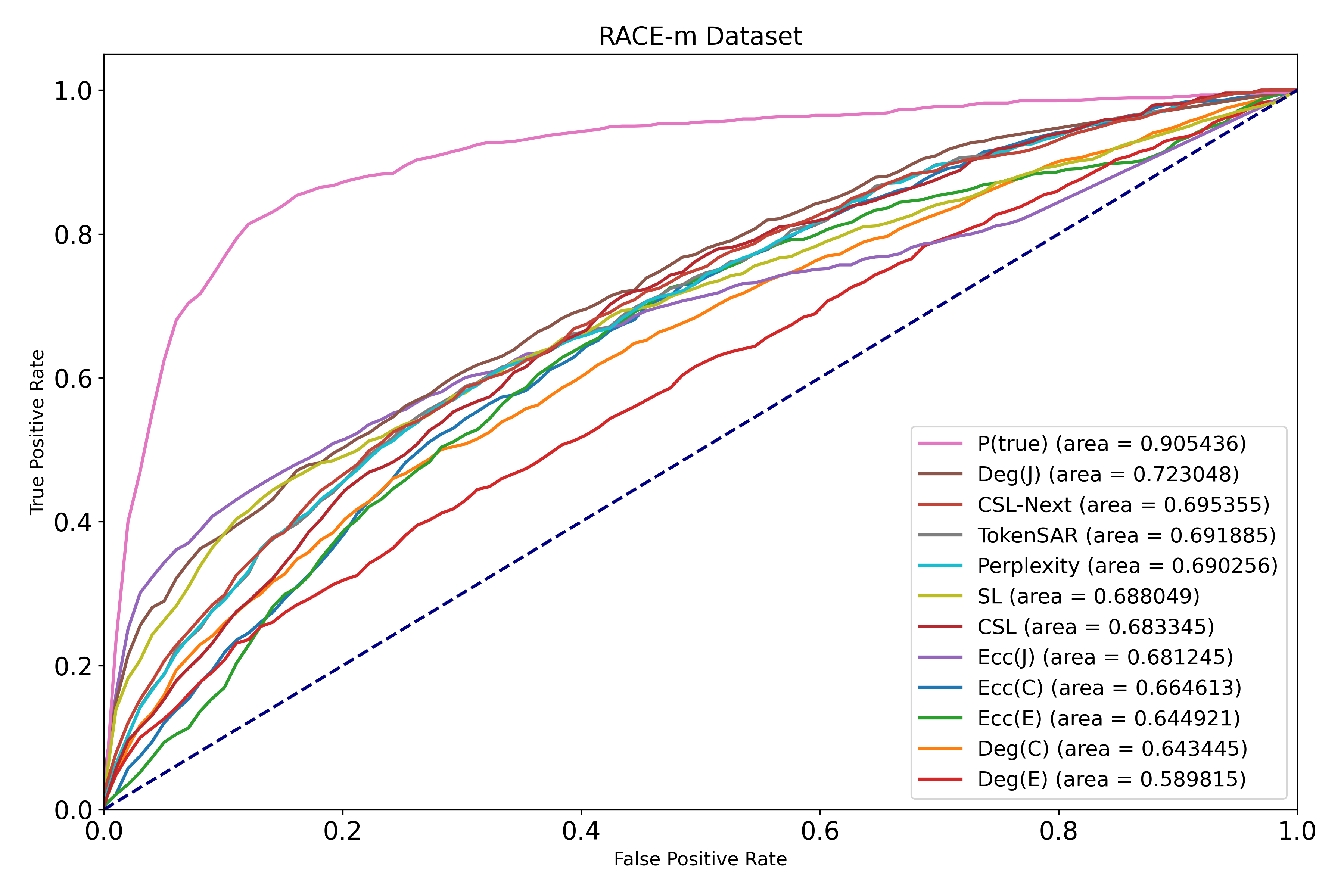}
        \caption{RACE-m Dataset}
        \label{fig:subfig3}
    \end{subfigure}
    \hfill
    \begin{subfigure}[b]{0.45\textwidth}
        \centering
        \includegraphics[width=\textwidth]{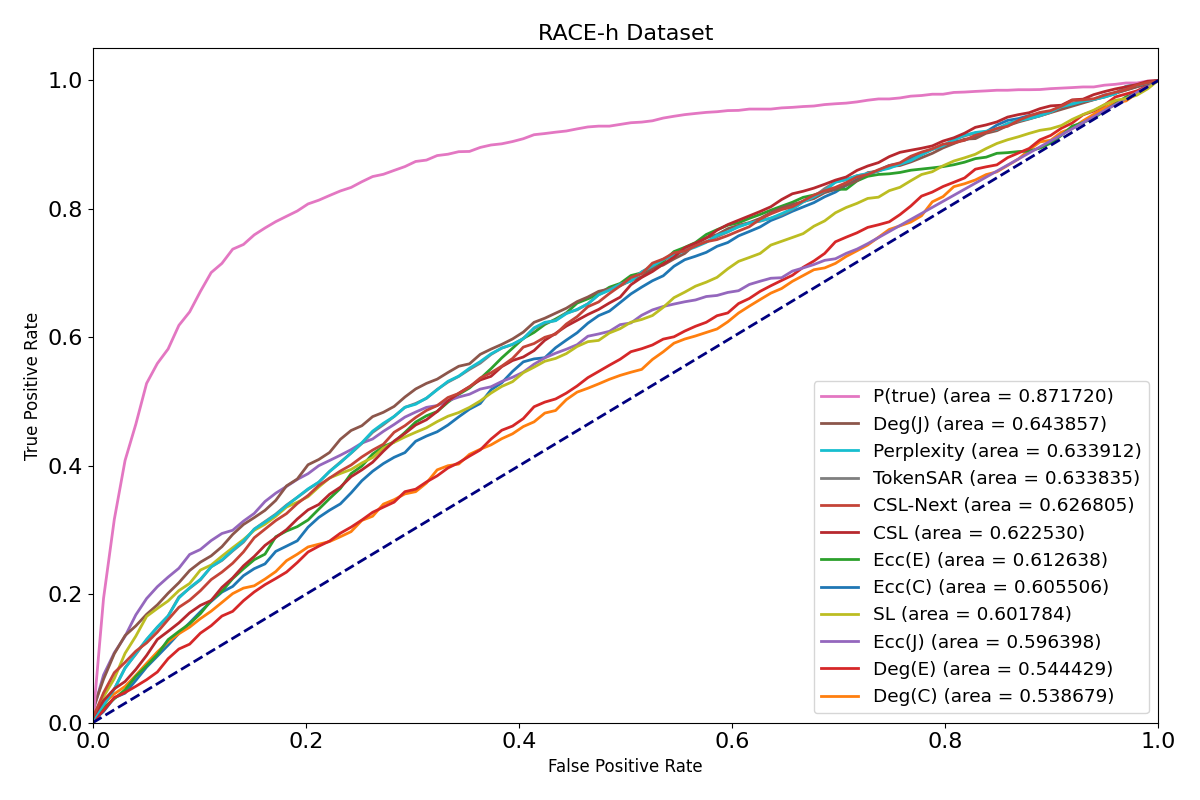}
        \caption{RACE-h Dataset}
        \label{fig:subfig4}
    \end{subfigure}

    \begin{subfigure}[b]{0.45\textwidth}
        \centering
        \includegraphics[width=\textwidth]{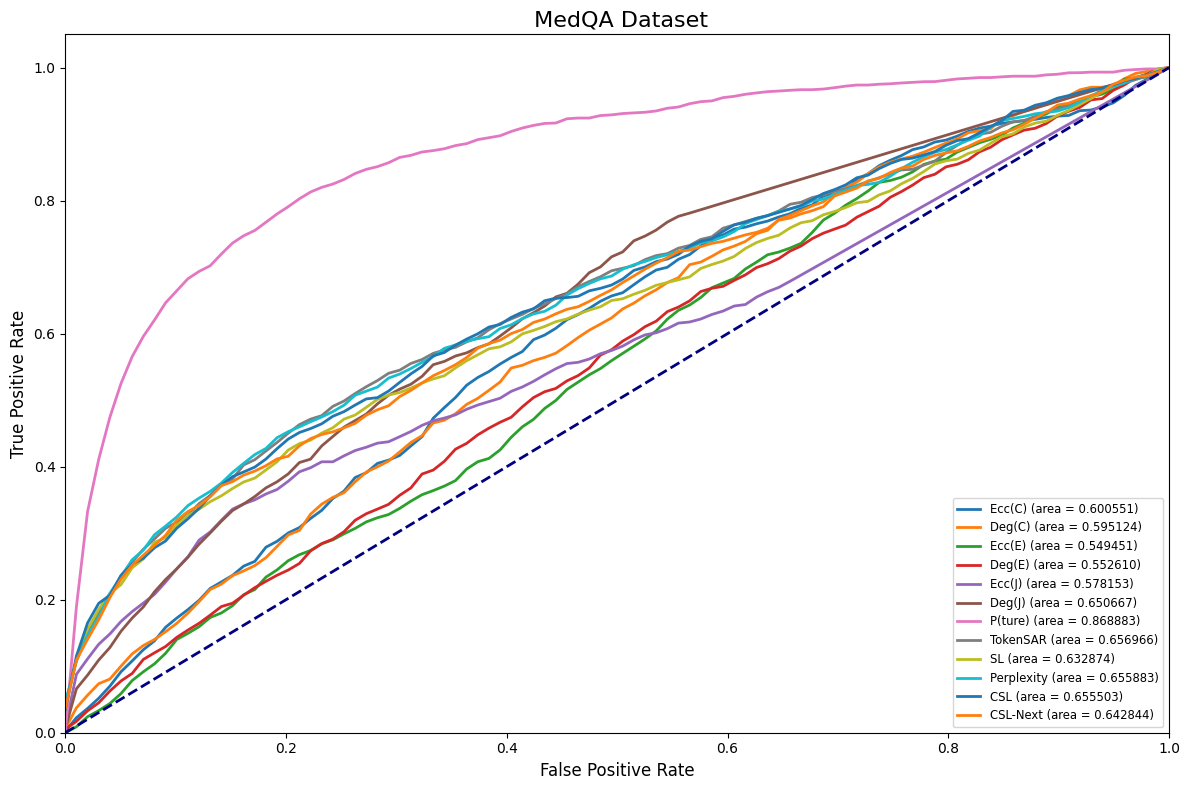}
        \caption{MedQA Dataset}
        \label{fig:subfig5}
    \end{subfigure}
    
    \caption{Comparison of different evaluation metrics using our method to quantify the \phiName model's confidence scores across five datasets (C-QA, QASC, RACE-m, RACE-h, MedQA), with increasing difficulty.}
    \label{appendix:fig:ROC}
\end{figure*}

\cref{appendix:fig:ROC} presents the ROC curves for \phiName.
\baselinePTrue achieves much better performance than other confidence measures on the more challenging datasets, likely because \phiName is a relatively advanced model.
On the easier datasets, where we could observe a bigger performance gap between different confidence measures, it is also interesting to see that the general shapes (and rankings at different FPR) are relatively consistent across C-QA and QASC, suggesting stability of \uqeval.

\def \FigAUROCVisHorizontalBar{
\begin{figure}[t]
  \centering
  \begin{subfigure}[b]{\columnwidth}
    \centering
    \includegraphics[width=\columnwidth]{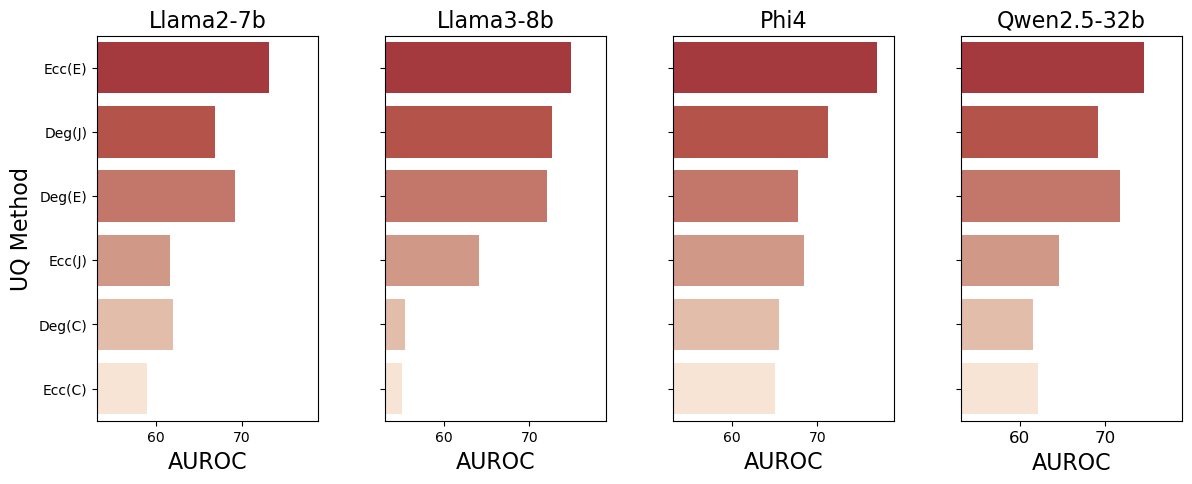}
    \caption{AUROC of different black-box methods.}
    \label{fig:cqa_blackbox}
  \end{subfigure}
  \begin{subfigure}[b]{\columnwidth}
    \centering
    \includegraphics[width=\columnwidth]{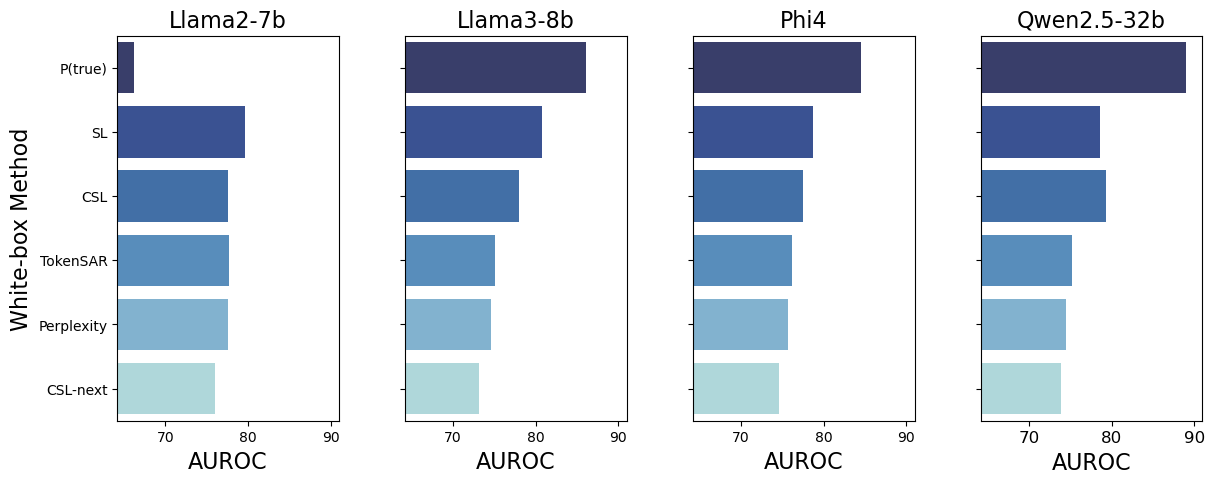}
    \caption{AUROC of different white-box methods.}
    \label{fig:another_dataset}
  \end{subfigure}

  \caption{(a) and (b) show the performance of 4 different LLMs and 12 different confidence estimation methods on the QASC dataset. A higher AUROC indicates better performance.}
  \label{fig:llm_perspective}
\end{figure}

\begin{figure}[t]
  \centering
  \begin{subfigure}[b]{\columnwidth}
    \centering
    \includegraphics[width=\columnwidth]{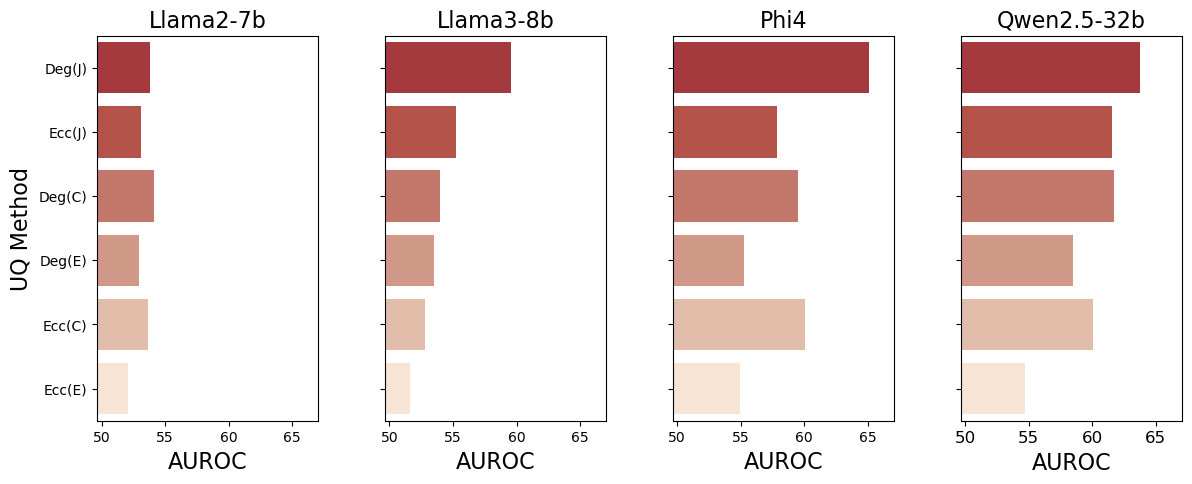}
    \caption{AUROC of different black-box methods.}
    \label{fig:cqa_blackbox}
  \end{subfigure}
  \begin{subfigure}[b]{\columnwidth}
    \centering
    \includegraphics[width=\columnwidth]{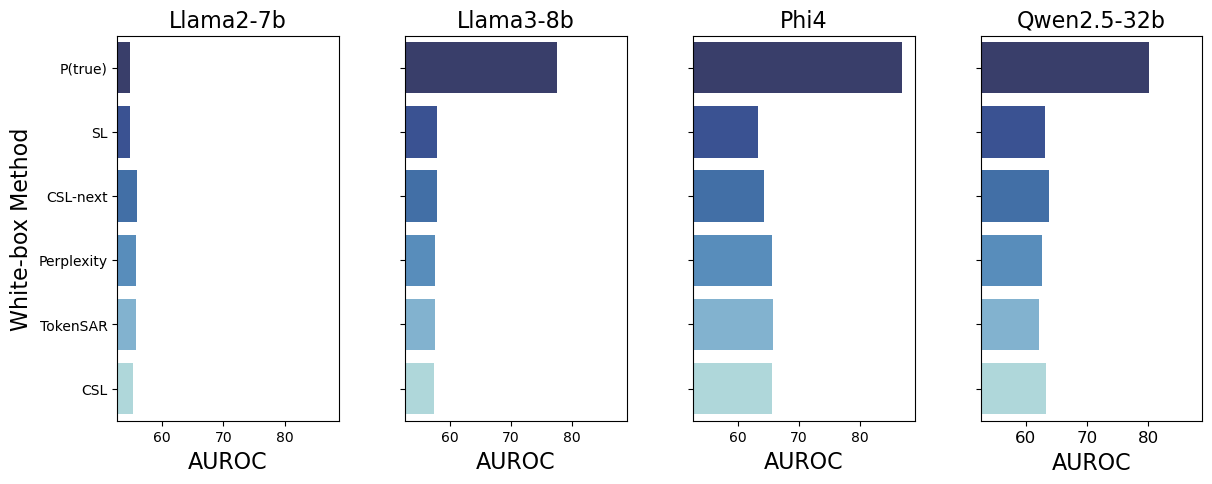}
    \caption{AUROC of different white-box methods.}
    \label{fig:another_dataset}
  \end{subfigure}

  \caption{(a) and (b) show the performance of 4 different LLMs and 12 different confidence estimation methods on the MedQA dataset. A higher AUROC indicates a better performance.}
  \label{fig:llm_perspective}
\end{figure}
}




